\def\year{2021}\relax
\documentclass[letterpaper]{article}
\usepackage{aaai21}
\usepackage{times}
\usepackage{helvet}
\usepackage{courier}
\usepackage[hyphens]{url}
\usepackage{graphicx}
\usepackage{natbib}
\usepackage{caption}
\usepackage{amsmath}
\usepackage{amssymb}
\usepackage{amsfonts}

\usepackage{algorithm,algcompatible}
\algnewcommand\INPUT{\item[\textbf{Input:}]}
\algnewcommand\OUTPUT{\item[\textbf{Output:}]}

\usepackage{multirow}

\usepackage{subcaption}
\usepackage{booktabs}
\usepackage{cancel}
\usepackage{hhline}
\usepackage{comment}

\usepackage[switch]{lineno}

\usepackage{xpatch}

\usepackage{makecell}
\newcommand{\Tstrut}{\rule{0pt}{2.6ex}}       
\newcommand{\Bstrut}{\rule[-1.2ex]{0pt}{0pt}} 
\newcommand{\TBstrut}{\Tstrut\Bstrut} 
\newcommand{\ngf}{\texttt{ngf}}
\newcommand{\ndf}{\texttt{ndf}}

\DeclareRobustCommand\onedot{\futurelet\@let@token\@onedot}
\def\onedot{. }
\def\eg{\emph{e.g}\onedot} \def\Eg.{\emph{E.g}\onedot}
 \def\Ie.{\emph{I.e}\onedot}

\def\etal{\emph{et al}\onedot}

\title{IB-GAN: Disentangled Representation Learning with \\ Information Bottleneck Generative Adversarial Networks}
\author {
    Insu Jeon, \textsuperscript{\rm 1}
    Wonkwang Lee, \textsuperscript{\rm 2}
    Myeongjang Pyeon, \textsuperscript{\rm 1}
    Gunhee Kim \textsuperscript{\rm 1} \\
}

\affiliations{
  \textsuperscript{\rm 1} Dept. of Computer Science and Engineering, Seoul National University, Republic of Korea (South)\\
  \textsuperscript{\rm 2} School of Computing, Korea Advanced Institute of Science and Technology, Republic of Korea (South)\\
  insuj3on@gmail.com, wonkwang.lee@kaist.ac.kr, mjpyeon@vision.snu.ac.kr, gunhee@snu.ac.kr\\
}

\begin{document}

\maketitle

\begin{abstract}
We propose a new GAN-based unsupervised model for disentangled representation learning. The new model is discovered in an attempt to utilize the Information Bottleneck (IB) framework to the optimization of GAN, thereby named IB-GAN. 
The architecture of IB-GAN is partially similar to that of InfoGAN but has a critical difference; an intermediate layer of the generator is leveraged to constrain the mutual information between the input and the generated output. The intermediate stochastic layer can serve as a learnable latent distribution that is trained with the generator jointly in an end-to-end fashion. As a result, the generator of IB-GAN can harness the latent space in a disentangled and interpretable manner. With the experiments on dSprites and Color-dSprites dataset, we demonstrate that IB-GAN achieves competitive disentanglement scores to those of state-of-the-art $\beta$-VAEs and outperforms InfoGAN. Moreover, the visual quality and the diversity of samples generated by IB-GAN are often better than those by $\beta$-VAEs and Info-GAN in terms of FID score on CelebA and 3D Chairs dataset.
\end{abstract}

\section{Introduction}
\label{sec:introduction}

Learning a good representation of data is one of the essential topics in machine learning research.
Although the \textit{goodness} of learned representation depends on the task, a general consensus on the useful properties of representation has been discussed through many recent studies \cite{bengio2013representation, Ridgeway:2016wp, Achille:2017tm}. 
A \textit{disentanglement}, one of such useful properties of representation, is often described as statistical independence of the data generative factors, which is semantically well aligned with human intuition (\emph{e.g.}, chair types or leg shapes on Chairs dataset \cite{Aubry14} and age or gender on CelebA dataset \cite{liu2015faceattributes}).
Distilling each important factor of data into a single independent direction of representation is hard to be done but invaluable for many other downstream tasks \cite{Ridgeway:2016wp,Achille:2017tm,Higgins:2017vt,Higgins:2017uo}. 

\begin{figure}[t!]
    \centering
    \includegraphics[height=3.4cm]{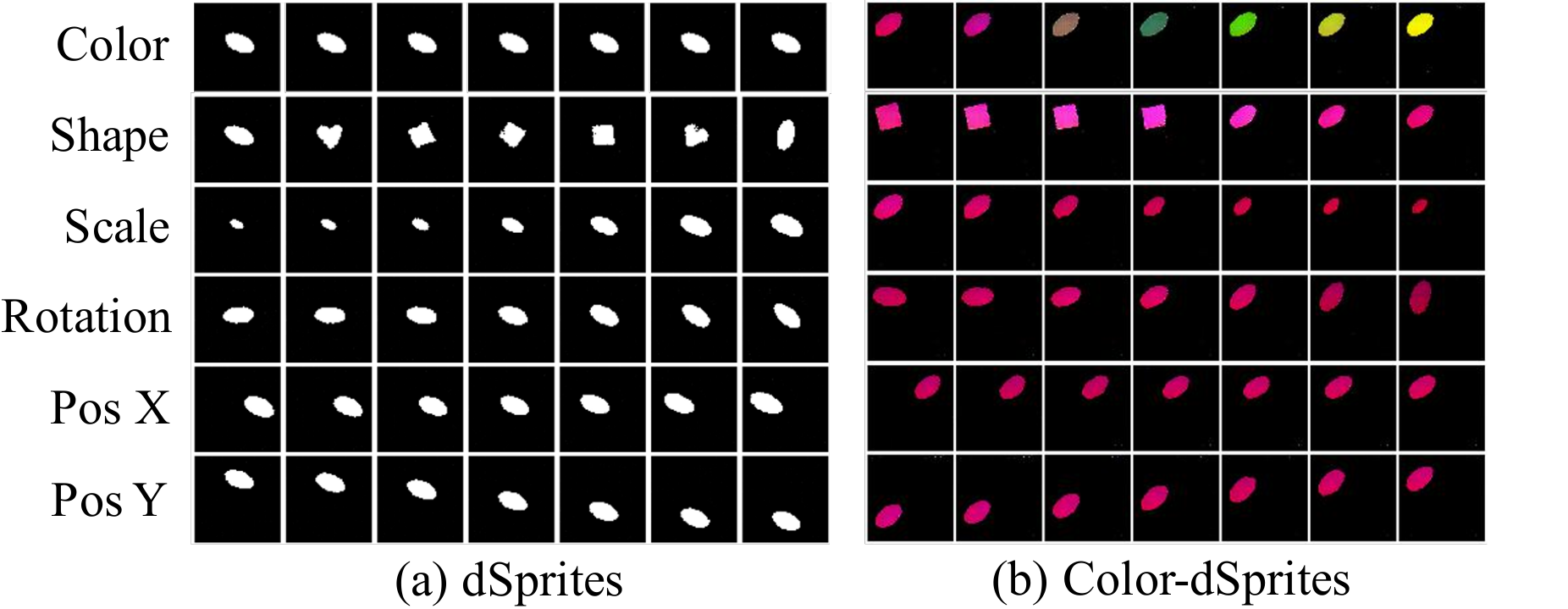}
    \vspace{-12pt}
    \captionof{figure}{Latent generative factors captured by IB-GAN on (a) dSprites and (b) Color-dSprites dataset. From top to bottom, each row corresponds to the factors of color, shape, scale, rotation, position Y, and position X.}
    \label{fig:dsprite}
    \vspace{-12pt}
\end{figure}

Recently, various models have been proposed for disentangled representation learning~\citep{hinton2011transforming, kingma2014semi, reed2014learning, Mathieu:2016df, siddharth2017learning, denton2017unsupervised, Jha:2018cc}. Despite their impressive results, they either require knowledge of ground-truth generative factors or weak-supervision (\emph{e.g.}, domain knowledge or partial labels). In contrast, among many unsupervised approaches~\citep{Dumoulin:2016td, Donahue:2016wo,Chen:2016tp,Higgins:2016vm,Burgess:2018uf,Kim:2018th,Chen:2018wu}, the two most popular approaches
maybe $\beta$-VAE \citep{Higgins:2016vm} and InfoGAN~\citep{Chen:2016tp}.

$\beta$-VAE \citep{Higgins:2016vm} demonstrates that encouraging the KL-divergence term of the Variational Autoencoder (VAE) objective \citep{Kingma:2013tz, Rezende:2014vm} by multiplying a constant $\beta>1$ induces strong statistical independence among the factors of latent representation. 
However, a high $\beta$ value can strengthen the KL regularization too much, leading to worse reconstruction fidelity than the standard VAE. 
On the other hand, InfoGAN~\citep{Chen:2016tp} is another fully unsupervised approach based on Generative Adversarial Network (GAN) \citep{Goodfellow:2014td}. It enforces the generator to learn disentangled representation by increasing the mutual information (MI) between the generated samples and the latent code. 
Although InfoGAN can learn independent factors well on relatively simple datasets such as MNIST \citep{lecun2010mnist}, it struggles to do so on more complicated datasets such as CelebA~\citep{liu2015faceattributes}  or 3D Chairs~\citep{Aubry14}. Reportedly, the disentangling performance of learned representation by InfoGAN is not as good as that of $\beta$-VAE and its variant models \citep{Higgins:2016vm, Kim:2018th, Chen:2018wu}.

Meanwhile, there have been many efforts \citep{Kim:2018th, Chen:2018wu,mathieu2019disentangling} to identify the mechanisms of disentanglement-promoting behavior in $\beta$-VAE. Based on the ELBO decomposition \citep{Hoffman:vz,Makhzani:2017ui}, it is revealed that the KL-divergence term in VAE can be factorized to the total correlation term \citep{Watanabe:1960cp,Hoffman:vz}, which essentially enforces the factorization of the marginal encoder and thus promotes the independence of learned representations in $\beta$-VAE. Besides, some other studies \citep{Burgess:2018uf, Alemi:2016tba, Alemi:2017va} identified that the KL regularization term in $\beta$-VAE corresponds to the mutual information (MI) minimization in the variational inference (VI) formulation \citep{Jordan:1999kv, Wainwright:2008du} of the Information Bottleneck (IB) theory \citep{Tishby:2000tq, Alemi:2016tba, Alemi:2017va}.

Based on the aforementioned studies, it is clear that the weakness of GAN-based disentangled representation learning comes from the fact that the model lacks any representation encoder or constraining mechanism for the representation.
In the conventional GAN model, a latent representation $z$ is sampled from a fixed latent distribution such as normal distribution, and the generator of GAN maps the whole normal distribution to the target images. Due to this, the latent representation $z$ can be utilized in a highly entangled way; an individual dimension of $z$ to not correspond well to the semantic features of data. 
Although InfoGAN supports an inverse mapping from data to latent code $c$, it still does not support disentangled relation as data is inverted to a fixed prior distribution. 

In this paper, we present a new GAN-based unsupervised model for disentangled representation learning. Specifically, a new GAN architecture is discovered in an attempt to solve the GAN's objective with IB framework, thereby named IB-GAN (\textit{Information Bottleneck GAN}).
The resulting architecture derived from variational inference (VI) formulation of the IB-GAN objective is partially similar to that of InfoGAN but has a critical difference; an intermediate layer of the generator is leveraged to constrain the mutual information between the input and the generated data.
The intermediate stochastic layer can serve as a learnable latent distribution that is trained with the generator jointly in an end-to-end fashion.
As a result, the generator of IB-GAN can harness the latent space in a disentangled and interpretable manner similar to $\beta$-VAE, while inheriting the merit of GANs (\emph{e.g.}, the model-free assumption on generators or decoders, producing good sample quality).

We summarize contributions of this work as follows:
\begin{enumerate}
\item
IB-GAN is a novel GAN-based model for unsupervised learning of disentangled representation. 
IB-GAN can be seen as an extension to the InfoGAN, supplementing an information constraining mechanism that InfoGAN lacks in the perspective of IB theory.
\item 
The resulting IB-GAN architecture derived from the variational inference (VI) formulation of the IB framework supports a trainable latent distribution via intermediate latent encoder between input and the generated data.
\item  With the experiments on dSprites \citep{Higgins:2016vm} and Color-dSprites dataset \citep{Burgess:2018uf, google1}, IB-GAN achieves competitive disentanglement scores to those of state-of-the-art $\beta$-VAEs \citep{Burgess:2018uf, Higgins:2016vm, Kim:2018th, Chen:2018wu} and outperforms InfoGAN \citep{Chen:2016tp}. The visual quality and diversity of samples generated by IB-GAN are often better than those by $\beta$-VAEs and InfoGAN on CelebA\citep{liu2015faceattributes} and 3DChairs \citep{Aubry14}. 
\end{enumerate}

\section{Background}
\label{sec:background} 

\subsection{InfoGAN: Information Maximizing GAN}
\label{sec:prelim_infogan}

Generative Adversarial Networks (GAN) \citep{Goodfellow:2014td} formulate a min-max adversarial game between two neural networks, a generator $G$ and a discriminator $D$: 
\begin{align} \label{eq:gan}
&\mathop{\min}_G \mathop{\max}_D \mathcal{L}_{\text{GAN}} (D,G) \nonumber \\ &= \mathbb{E}_{p(x)}[\log(D(x))] + \mathbb{E}_{p(z)} [\log (1-D(G(z))]. 
\end{align}
The discriminator $D$ aims to distinguish well between real samples $x \sim p(x)$ and synthetic samples created by the generator $G(z)$ with random noise $z \sim p(z)$, while the generator $G$ is trained to produce realistic sample that is indistinguishable from true sample. 
Under an optimal discriminator $D^*$, Eq.(\ref{eq:gan}) theoretically minimizes the Jensen-Shannon divergence between the synthetic and the true sample distribution: $JS(G(z)||p(x))$. 
However, Eq.(\ref{eq:gan}) does not have any specific guidance on how $G$ utilizes a mapping from $z$ to $x$. That is, the variation of $z$ in any independent dimension often yields entangled effects on a generated sample $x$.

InfoGAN \citep{Chen:2016tp} is capable of learning disentangled representations without any supervision. 
To do so, 
the objective of 
InfoGAN accommodates
a mutual information maximization term between an additional latent code $c$ and a generated sample $x=G(z,c)$: 
\begin{align}
\label{eq:infogan}
&\min_{G}\max_D \mathcal{L}_{\text{InfoGAN}}(D,G) \nonumber \\
&= \mathcal{L}_{\text{GAN}}(D,G) - I(c,G(z,c)), 
\end{align}
where $I(\cdot,\cdot)$ denote MI. Also, $c$ and $z$ are latent code and not interpretable (or in-compressible) noise respectively. To optimize Eq.(\ref{eq:infogan}), the variational lower-bound of MI is exploited similarly to the IM algorithm \citep{Barber:2003uw}.

\subsection{Information Bottleneck Principle}
\label{sec:prelim_ib}

Let the input variable $X$ and the target variable $Y$ distributed according to some joint data distribution $p(x,y)$. 
The goal of IB \citep{Tishby:2000tq, Alemi:2016tba, Alemi:2017va, peng2018variational} is to obtain a compressive representation $Z$ from the input $X$, while maintaining the predictive information about the target $Y$ as much as possible. The objective for the IB is 
\begin{align} \label{eq:obj_ib}
\max_{q_\phi(z|x)} \mathcal{L}_{\text{IB}} = I(Z,Y) - \beta I(Z,X),
\end{align}
where $I(\cdot,\cdot)$ denotes MI and $\beta \ge 0$ is a Lagrange multiplier. 
Therefore, IB aims at obtaining the optimal representation encoder\footnote{$\phi$ is the parameter of representation encoder model.} $q_\phi(z|x)$ 
that simultaneously balances the tradeoff between the maximization and minimization of both MI terms. Accordingly, the learned representation $Z$ can act as a minimal sufficient statistic of  $X$ for predicting $Y$. The IB principle provides an intuitive meaning for the \textit{good representation} from the perspective of information theory.
Recent studies \cite{Burgess:2018uf,Achille:2017tm, zhao2018information} show that the variational inference (VI) formulation \citep{Jordan:1999kv, Wainwright:2008du} of IB Eq.\eqref{eq:obj_ib} is equivalent to the objective of $\beta$-VAE when the task is self-reconstruction (\emph{e.g.}, $Y=X$).

\begin{figure}[t!]
    \centering
    \includegraphics[height=3.8cm]{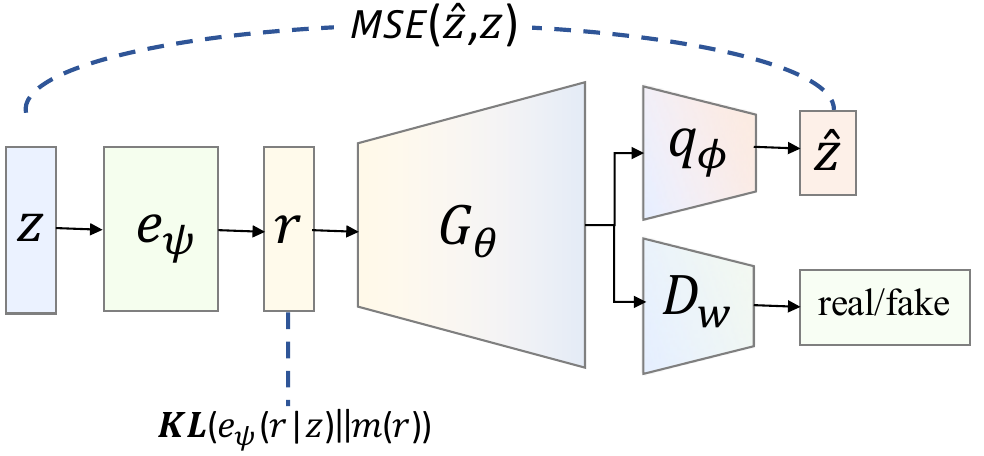}   
    \caption{An illustration of IB-GAN. It adopts a representation encoder $e_\psi(r|z)$ and a KL-divergence loss derived from IB theory. Since the encoder $q_\phi(r|z)$ is assumed as Gaussian, it is convenient to define $m(r)$ as Gaussian. The MSE loss is computed by the decoder $\log q_\phi(z|x)$ in Eq.\eqref{eq:lower_ibgan}. }
        \label{fig:model}
\end{figure}

\section{IB-GAN: Information Bottleneck GAN}
\label{sec:ibgan} 

The motivation for IB-GAN is straight forwards. We can identify that InfoGAN's objective in Eq.\eqref{eq:infogan} lacks a MI minimization term compare to the IB objective in Eq.\eqref{eq:obj_ib}. Thus, we utilize the MI minimization term to InfoGAN's objective to get the IB-GAN objective as follows: 
\begin{align} \label{eq:ibgan2}
&\min_{G}\max_D \mathcal{L}_{\text{IB-GAN}}(D,G) \nonumber \\
&= \mathcal{L}_{\text{GAN}}(D,G) - \big[ I^L(z,G(z)) - \beta I^U(z,G(z)) \big],\\
&\;\;\;\;\;\text{s.t.} \;\; I^L(z,G(z)) \le I_g(z,G(z)) \le I^U(z,G(z)),  \label{eq:ibgan}
\end{align}
where $I^L(\cdot,\cdot)$ and $I^U(\cdot,\cdot)$ denote the lower and upper-bound of \textit{generative} MI\footnote{The \textit{generative} mutual information (MI) is described as $I_g(Z,X)=E_{p_\theta(x|z)p(z)}[p_\theta(x|z)p(z)/p_\theta(x)p(z)]$. This formulation of MI is also exhibited in InfoGAN and IM algorithm \citep{Chen:2016tp, Barber:2003uw}.} respectively. One important change in Eq.(\ref{eq:ibgan2}) compared to Eq.(\ref{eq:infogan}) is adopting the upper-bound of MI with a trade-off coefficient $\beta$, analogously to that of $\beta$-VAE and the IB objective\footnote{The incompressible noise variable $z$ is not necessarily required for modeling InfoGAN \citep{Srivastava:2017tt, zhao2018information}. So, we can omit the incompressible noise $z$. Here, $z$ has the same role as the latent code $c$ in InfoGAN.}. More discussion regarding this parameter is presented in the next section, and the effect is explored in the experimental section. 

\subsection{Optimization of IB-GAN}
\label{sec:opt_ibgan} 

For the optimization of Eq.\eqref{eq:ibgan2}, we first define a tractable lower-bound of the \textit{generative} MI in Eq.(\ref{eq:ibgan}) using the similar derivation exhibited in \citep{Chen:2016tp,Barber:2006,alemi2018gilbo}. 
For the brevity, we use probabilistic model notion (\emph{i.e.}, $p_\theta(x|z)=\mathcal{N}(G_\theta(z), \mathbf{1})$ for the generator. Then, the variational lower-bound $I^L(z,G(z))$ of the \textit{generative} MI in Eq.(\ref{eq:ibgan}) is given as
\begin{align}\label{eq:lowerbound}
&I_g(z,G(z))= \mathbb{E}_{p_\theta(x|z)p(z)} [\log \frac{p_\theta(x|z)p(z)}{p_\theta(x)p(z)}] \nonumber \\
&\geq ~I^L(z,G(z)) = \mathbb{E}_{p_\theta(x|z)p(z)}[\log\frac{q_\phi(z|x)}{p(z)}] 
\\
&=\mathbb{E}_{p_\theta(x|z)p(z)}[\log q_\phi(z|x)] + H(z).\label{eq:lowerbound3}
\end{align}
In Eq.(\ref{eq:lowerbound}), the lower-bound holds thanks to positivity of KL-divergence. A variational reconstructor $q_\phi(z|x)$ is introduced to approximate the quantity $p_\theta(z|x)=p_\theta(x|z)p(z)/p_\theta(x)$.
Intuitively, by improving the reconstruction of an input code $z$ from a generated sample $x=G(z)$ using the $q_\phi(z|x)$, we can promote the statistical dependency between the generator $G(z)$ and the input latent code $z$ \citep{Chen:2016tp, Barber:2003uw}.

\makeatletter
\xpatchcmd{\algorithmic}
  {\ALG@tlm\z@}{\leftmargin\z@\ALG@tlm\z@}
  {}{}
\makeatother

\begin{algorithm}[t]
  \caption{IB-GAN training algorithm} \label{alg:ib-gan}
  \begin{algorithmic}[]
    \State \textbf{Input:} batch size $B$, hyperparameters $\beta$, and the learning rates $\eta_\phi$, $\eta_\theta$, $\eta_\psi$, and $\eta_w$ of the parameter of reconstructor, generator, encoder, and discriminator model respectively.
    \WHILE{not converged}
    \STATE Sample $\left\{z^1, \dots, z^B\right\} \sim p(z)$
    \STATE Sample $\left\{x^1, \dots, x^B\right\} \sim p(x)$
    \STATE Sample $\left\{r^1, \dots, r^B\right\} \sim e_\psi(r|z^i)$ for $i\in\left\{ 1 \dots B \right\}$
    \STATE Sample $\left\{x^1_g, \dots, x^B_g\right\} \sim p_\theta(x|r^i)$ for $i\in\left\{ 1 \dots B \right\}$
    \STATE Sample $\left\{\hat{z}^1, \dots, \hat{z}^B\right\} \sim q_\phi(z|x^i_g)$ for $i\in\left\{ 1 \dots B \right\}$
    
    \STATE $g_{\phi} \leftarrow \nabla_{\phi}\frac{1}{B}\sum_i {(\hat{z}^i - z^i)^2}$
    \STATE $g_{w} \leftarrow -\nabla_{w}\frac{1}{B}\sum_{i}{\log\sigma(D_{w}(x_g^i))+ \log(1-\sigma(D_{w}(x^i))}$
    \STATE $g_{\theta} \leftarrow \nabla_{\theta}\frac{1}{B}\sum_{i}{\log\sigma(D_{w}(x_g^i))} - {(\hat{z}^i - z^i)^2}$
    \raggedright
    \STATE $g_{\psi} \leftarrow \nabla_{\psi}\frac{1}{B}\sum_{i}{\log\sigma(D_{w}(x_g^i))}- {(\hat{z}^i - z^i)^2} + {\beta KL(e_\psi(r|z^i)||m(r))}$
    \STATE $\phi \leftarrow \phi-\eta_\phi g_\phi;~ w \leftarrow w-\eta_w g_w;$
    \STATE $~ \theta \leftarrow \theta-\eta_\theta g_\theta;~ \psi \leftarrow \psi-\eta_\psi g_\psi$
    \ENDWHILE
  \end{algorithmic}
  \vspace{-2pt}
\end{algorithm}

In contrast to the lower-bound, obtaining a practical variational upper-bound of the \textit{generative} MI in Eq.(\ref{eq:ibgan}) is not trivial. If we follow the similar approach discussed in previous studies~\citep{Alemi:2016tba, Alemi:2017va}, the variational upper-bound $I^U(z,G(z))$ of the generative MI is derived as
\begin{align}\label{eq:badupperbound}
&~I_g(z,G(z)) = \mathbb{E}_{p_\theta(x|z)p(z)} [\log \frac{p_\theta(x|z)\bcancel{p(z)}}{p_\theta(x)\bcancel{p(z)}}] \nonumber\\
&\le ~I^U(z,G(z)) = \mathbb{E}_{p_\theta(x|z)p(z)} \log[\frac{p_\theta(x|z)}{d(x)}],
\end{align}
where $d(x)$ approximates the generator marginal $p_\theta(x)=\sum_z p_\theta(x|z)p(z)$. 
In theory, we can choose any approximation model $d(x)$. However, one important concern here is that it is difficult to correctly identify a proper approximation model for $d(x)$ in practice. Given that the upper-bound $I^U(z,G(z))$ is identical to $KL(p_\theta(x|z)||d(x))$ in Eq.(\ref{eq:badupperbound}), $d(x)$ acts as an image prior. Thus, any improper choice of $d(x)$ may severely downgrade the quality of synthesized samples from the generator $p_\theta(x|z)$.
We might mitigate this by introducing another GAN loss for the KL divergence, but the effective prior model $d(x)$ is still missing. 

For this reason, we propose another formulation of the variational upper-bound on the \textit{generative} MI, inspired by the recent studies of deep-learning with IB principle \citep{Tishby:2015cj, Achille:2017tm, Achille:ej}. We define an additional stochastic model $e_\psi(r|z)$ that takes a noise input vector $z$ and produces an intermediate stochastic representation $r$. In other words, we let $x=G(r(z))$ instead of $x=G(z)$; then we can express the generator\footnote{In this case, we let $p_\theta(x|r)=\mathcal{N}(G_\theta(r),\mathbf{1})$.} as $p_\theta(x|z)=\sum_r p_\theta(x|r)e_\psi(r|z)$. 
Subsequently, a new variational upper-bound $I^U(z,R(z))$ can be obtained as
\begin{align} 
&~I_g(z,G(R(z))) \nonumber \\
&\leq I(z,R(z))
= \mathbb{E}_{e_\psi(r|z)p(z)} [ \log  \frac{e_\psi(r|z)\bcancel{p(z)}}{e_\psi(r)\bcancel{p(z)}}] \label{eq:upperbound}\\
&\leq I^U(z,R(z)) = \mathbb{E}_{e_\psi(r|z)p(z)} \log[\frac{e_\psi(r|z)}{m(r)}]. \label{eq:upperbound2}
\end{align}
The first inequality in Eq.(\ref{eq:upperbound}) holds due to the Markov property \citep{Tishby:2015cj}: if any generative process follows $Z \rightarrow R \rightarrow X$, then $I(Z,X) \leq I(Z,R)$. The second inequality in Eq.(\ref{eq:upperbound2}) holds from the positivity of KL divergence. Any model for $m(r)$ can be flexibly used to approximate the representation marginal $e_\psi(r)$ (\emph{e.g.}, Gaussian).
This new formulation of the variational upper-bound in Eq.(\ref{eq:upperbound2}) allows us to constrain the \textit{generative} MI without directly affecting the generator
in Eq.(\ref{eq:badupperbound}) via the intermediate representation encoder model $e_\psi(r|z)$ and the prior $m(r)$.

Finally, from the lower-bound in Eq.(\ref{eq:lowerbound3}) and the upper-bound in Eq.(\ref{eq:upperbound2}), a variational approximation of Eq.(\ref{eq:ibgan2}) can be obtained as
\begin{align} \label{eq:lower_ibgan}
&\min_{G,q_\phi,e_\psi}\max_D \; \mathcal{\Tilde{L}}_{\text{IB-GAN}}(D,G,q_\phi,e_\psi) \nonumber\\
&=\mathcal{L}_{\text{GAN}}(D,G) - \big( \mathbb{E}_{p(z)}[\mathbb{E}_{p_\theta(x|r)e_\psi(r|z)}[\log q_\phi(z|x)] \nonumber\\
&\;\;\;-\beta \text{KL}(e_\psi(r|z)||m(r))]\big).
\end{align}
We define the encoder $e_\psi(r|z)$ as a stochastic model $N(\mu_\psi(z),\sigma_\psi(z))$ and the prior $m(r)$ as $N(\mathbf{0},\mathbf{1})$, as done in VAEs \citep{Kingma:2013tz}. 
The optimization of Eq.(\ref{eq:lower_ibgan}) can be done by alternatively updating the parameters of the generator $p_\theta(x|r)$, the representation encoder $e_\psi(r|z)$, the variational reconstructor $q_\phi(z|x)$ and the discriminator $D$ using any SGD-based algorithm. A reparameterization trick~\citep{Kingma:2013tz} is employed to backpropagate gradient signals to the stochastic encoder. The overall architecture of IB-GAN is presented in Figure \ref{fig:model}, and the training procedure is described in Algorithm 1.

\subsection{Related Work and Discussion}
\label{sec:discuss_ibgan}

\subsubsection{Connection to rate-distortion theory.} 
IB theory is a generalization of the Rate-Distortion (RD) theory \citep{Tishby:2000tq}, in which the rate $\mathcal{R}$ is the code length per data sample to be transmitted through a noisy channel, and the distortion $\mathcal{D}$ represents the approximation error of reconstructing the input from the source code \citep{Alemi:2017va,shannon1948mathematical}. The goal of RD-theory is to minimize $\mathcal{D}$ without exceeding a certain level of rate $\mathcal{R}$, formulated as $\min_{\mathcal{R},\mathcal{D}} \mathcal{D} + \beta \mathcal{R}$, where $\beta \in [0, \infty]$ decides a theoretically achievable optimum in the auto-encoding limit \citep{Alemi:2017va}.

IB-GAN can be described in terms of the RD-theory. Here, the goal is to deliver an input code $z$ through a noisy channel (\emph{i.e.}, deep neural networks). Both $r$ and $x$ are regarded as the encoding of input $z$. The distortion in IB-GAN corresponds to the reconstruction error of the input $z$ estimated from the variational encoder $q_\phi(z|x(r))$. 
          
The rate $\mathcal{R}$ of the intermediate representation $r$ is related to $KL(e_\psi(r|z)||m(r))$, which measures the inefficiency (or the excess rate) of the representation encoder $e_\psi(r|z)$ in terms of how much it deviates from the approximating representation prior $m(r)$. 
Hence, $\beta$ in Eq.\eqref{eq:lower_ibgan} controls the compressing level of the information contained in $r$ for reconstructing input $z$. It constrains the amount of shared information between the input code $z$ and the output image by the generator $x=G(r(z))$ without directly regularizing the output image itself.
In addition, the GAN loss $\mathcal{L}_{\text{GAN}}$ in Eq.\eqref{eq:lower_ibgan} can be understood as a rate constraint of the image in the context of RD-theory since the GAN loss approximates $JS(G(z)||p(x))$ \citep{Goodfellow:2014td} between the generator and the empirical data distribution $p(x)$.

\subsubsection{Comparison between IB-GAN and $\beta$-VAE.} The resulting architecture of IB-GAN is partly analogous to that of $\beta$-VAE since both are derived from the IB theory\footnote{IB-GAN's objective is derived from the \textit{generative} MI, while $\beta$-VAE's objective is derived from the \textit{representational} MI in \cite{Alemi:2016tba, Alemi:2017va}.}.
$\beta$-VAE tends to generate blurry output images due to large $\beta$ \cite{Kim:2018th, Chen:2018wu}. Setting $\beta$ to large value minimizes the excess rate of encoding $z$ in $\beta$-VAE, but this also increases the reconstruction error (or the distortion) \cite{Alemi:2017va}. 
In contrast, IB-GAN may not directly suffer from this shortcoming of $\beta$-VAE. The generator of IB-GAN learns to encode image $x$ by minimizing the rate (\emph{i.e.}, $JS(G(r)||p(x))$) inheriting the merit of InfoGANs (\emph{e.g.}, an implicit decoder model can be trained to produce a good quality of images).
One possible drawback of IB-GAN architecture is that it does not directly map the representation encoder to output $r$ from the real image $x$: $q(r|x)$. To obtain the representation $r$ back from the real data $x$, we need a two-step procedure: sampling $z$ from the learned reconstructor $q_\phi(z|x)$ and putting it to the representation encoder $e_\psi(r|z)$. However, the latent representation $r$ obtained from this procedure is quite compatible with those of $\beta$-VAEs as we will see in the experimental section.

\subsubsection{Disentanglement-promoting behavior of IB-GAN.} The disentanglement-promoting behavior of $\beta$-VAE is encouraged by the KL divergence. Since the prior distribution is often assumed as a fully factored Gaussian distribution, the KL divergence term in VAE objective can be decomposed into the form containing a total correlation (TC) term \citep{Watanabe:1960cp,Hoffman:vz}, which essentially enforces the statistical factorization of the representation \citep{Kim:2018th, Chen:2018wu, Burgess:2018uf}. In IB-GAN, a noise $z$ is treated as the input source instead of image $x$. Therefore, the disentangling mechanism of IB-GAN is slightly different from that of $\beta$-VAE.

The disentanglement-promoting behavior of IB-GAN can be described in term of the RD-theory as follow:
(1) The efficient encoding scheme for the (intermediate) latent representation $r$ can be learned by minimizing $KL(e_\psi(r|z)||m(r))$ with a factored Gaussian prior $m(r)$, which promotes statistical factorization of the coding $r$ similar to that of VAE. (2) The efficient encoding scheme for $x$ is defined by minimizing the divergence between $G(z)$ and the data distribution $p(x)$ via the discriminator, which promotes the encoding of $x$ to be a realistic image. 
(3) Maximizing $I^L(z,G(z))$ in IB-GAN indirectly maximizes $I(r,G(r))$ too since $I(z,G(z)) \leq I(r,G(r))$ from the Markov propriety \citep{Tishby:2015cj}. That is, maximizing the lower-bound of MI increases the statistical dependency between the coding $r$ and $x=G(r)$, while both encoding $r$ and $x$ need to be efficient in terms of their rates (\emph{e.g.}, the upper-bound of MI and the GAN loss). Therefore, an independent directional change in the representation encoding $r$ tends to be well aligned with 
a predominant factor of variation in the image $x$.

\subsubsection{Other characterizations of IB-GAN.}
IB-GAN can softly constrain the \textit{generative} MI by the variational upper-bound derived in Eq.\eqref{eq:upperbound2}. 
In this regard, the variational encoder of IB-GAN can be seen as a hierarchical trainable prior for the generator. If $\beta$ in Eq.\eqref{eq:lower_ibgan} is zero, the IB-GAN objective reduces to that of InfoGAN. In contrast, if $\beta$ is too large such that the KL-divergence term is almost zero, then there would be no difference between the samples from the representation encoder $e_\psi(r|z)$ and the distortion prior $m(r)$. Then, both representation $r$ and generated data $x$ contain no information about $z$ at all, resulting in the signal from the reconstructor being meaningless to the generator. If we further exclude the lower-bound of MI in Eq.\eqref{eq:lower_ibgan}, the IB-GAN objective reduces to that of vanilla GAN with an input $r\sim m(r)$.

\begin{figure*}[ht!]
    \begin{subfigure}[b]{0.31\textwidth}
    \includegraphics[width=1\linewidth]{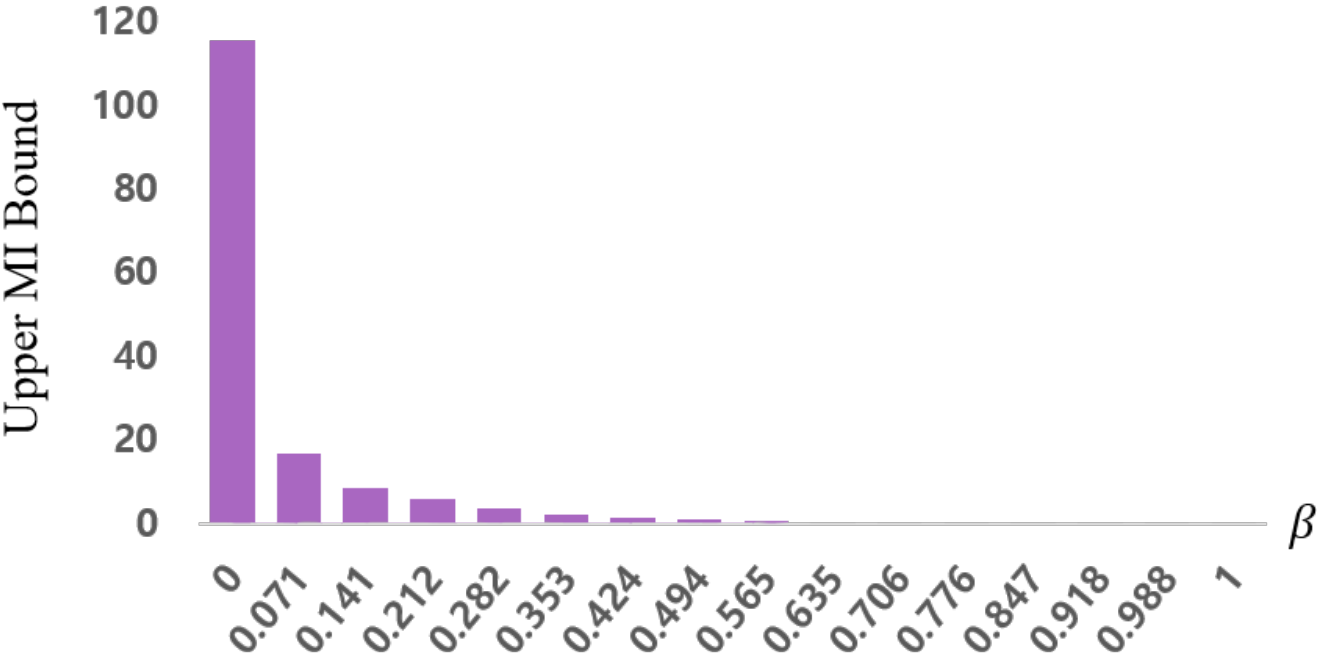}
        \caption[]
        {{\small $I^U(z,G(z))$ vs $\beta$.}}    
    \end{subfigure}
    \hfill \hspace{-4pt}
    \begin{subfigure}[b]{0.31\textwidth}
    \includegraphics[width=1\linewidth]{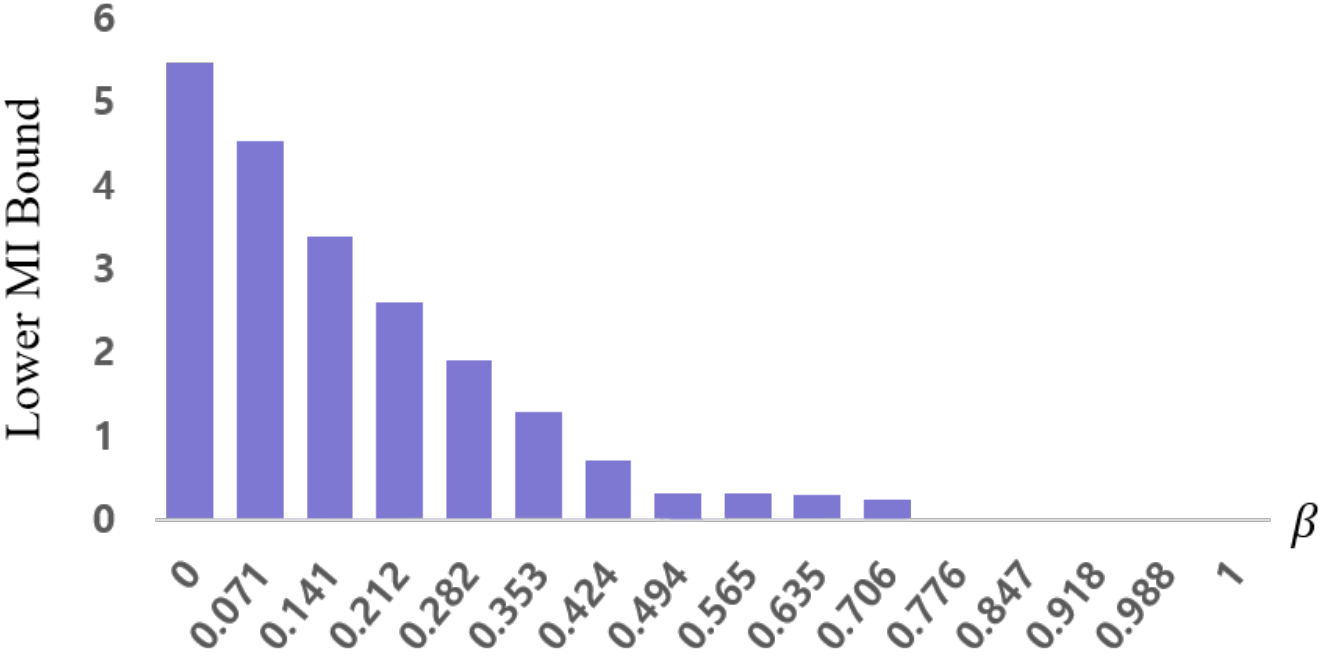}
        \caption[]
        {{\small $I^L(z,G(z))$ vs $\beta$.}}    
    \end{subfigure}
    \hfill \hspace{-4pt}
    \begin{subfigure}[b]{0.31\textwidth}
    \includegraphics[width=1\linewidth]{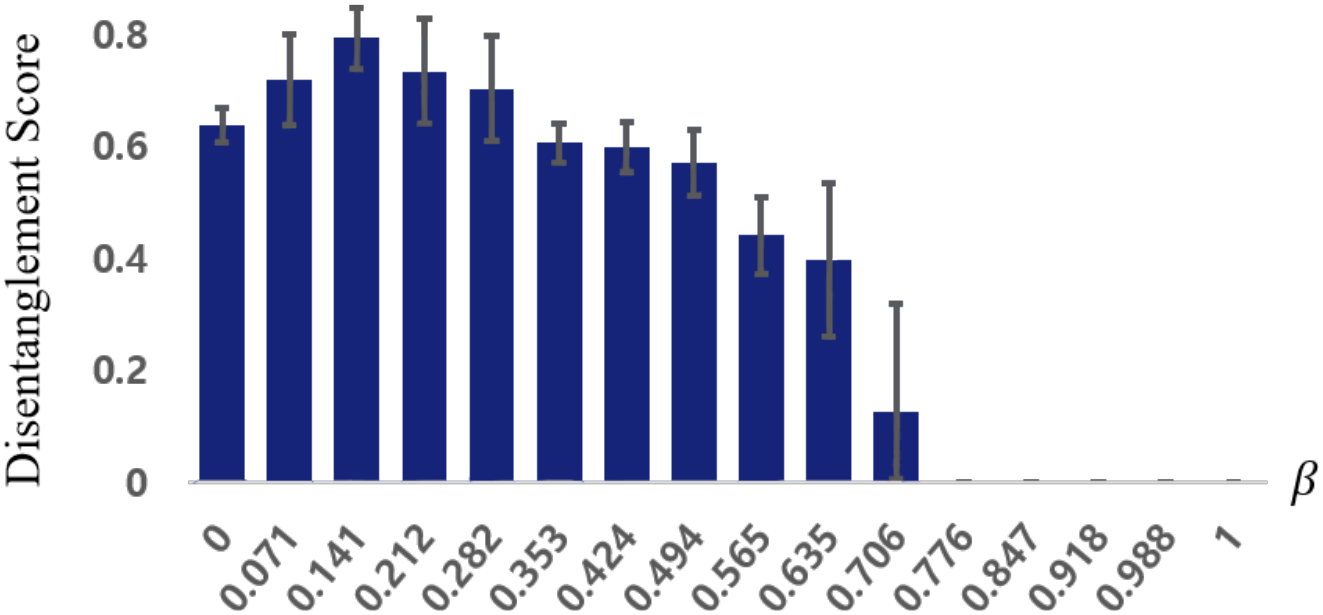}
        \caption[]
        {{\small Disentanglement score vs $\beta$.}}
    \end{subfigure}
    \caption[ ]
        {Effects of $\beta$ on the converged upper/lower-bound of MI and disentanglement metric scores~\cite{Kim:2018th}.}
    \label{dsprite-beta-mig}
\end{figure*}

\subsubsection{Variational bounds on generative MI.} Maximizing the variational lower-bound of \textit{generative} MI has been employed in IM algorithm \citep{Barber:2006} and InfoGAN \citep{Chen:2016tp}. Recently, Alemi and Fischer \cite{alemi2018gilbo} propose the lower-bound of \textit{generative} MI, named GILBO, as a data-independent measure that can quantify the complexity of the learned representations for trained generative models. They discover that the lower-bound is correlated with the image quality metrics of generative models such as INCEPTION~\citep{barratt2018note} and FID \cite{heusel2017gans} scores. On the other hand, we propose a new approach of upper-bounding the \textit{generative} MI, based on the causal relationship of deep learning architecture, and show the effectiveness of the upper-bound by measuring the disentanglement scores~\citep{Kim:2018th} on the learned representation.

\begin{figure}[t!]
        \centering
         \includegraphics[width=8.4cm]{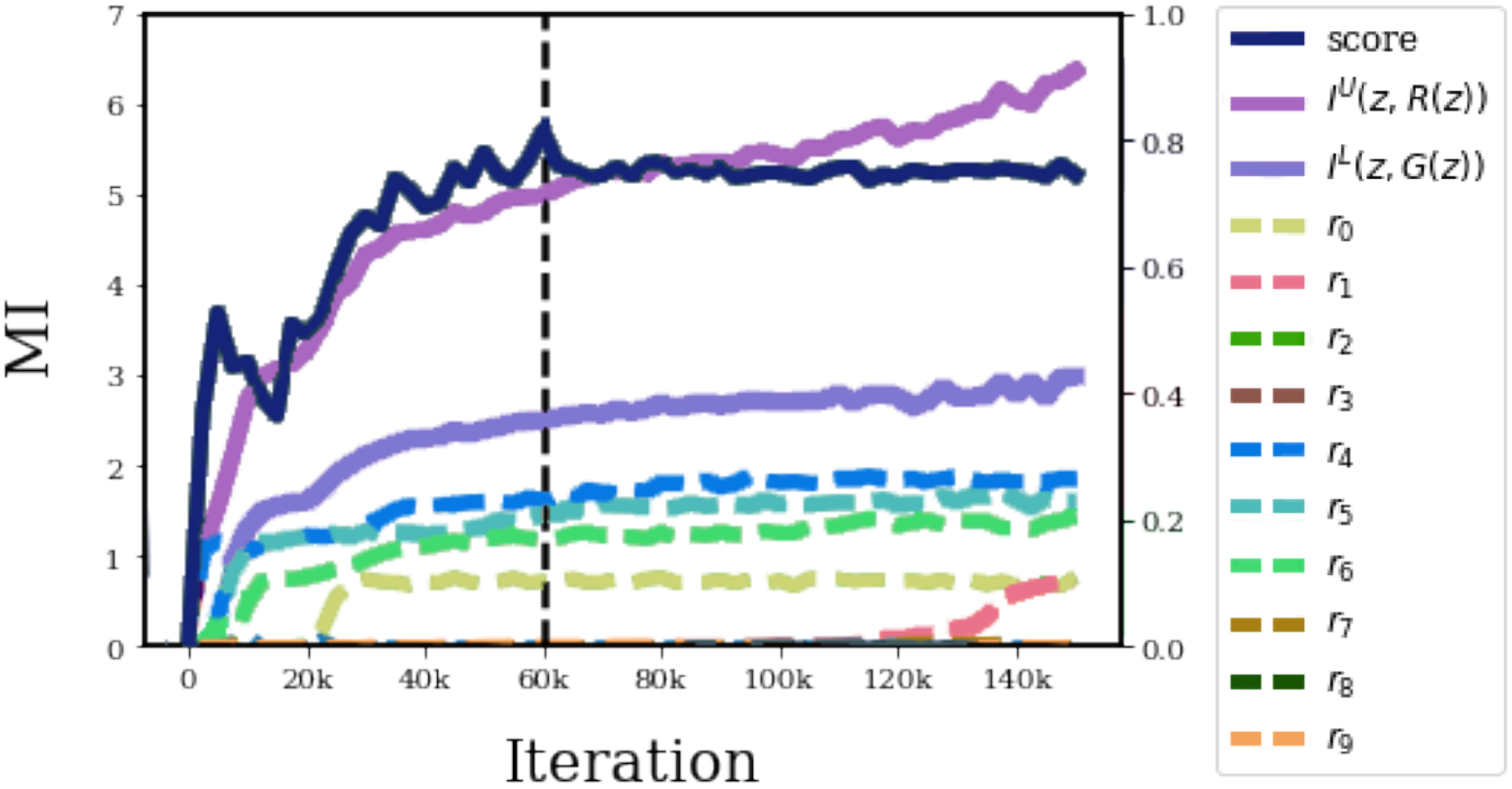}
        \captionof{figure}{ The plot of variational upper-bound and lower-bound of MI with independent $\mbox{KL}(e(r_i|z)||m(r_i))$ values for all $r_i$~($i=1,\ldots,10$) and disentanglement scores~\cite{Kim:2018th} over 150K training iterations.
The vertical dashed black line represents the iteration at the highest disentanglement score.}
        \label{fig:convergence}
    \hspace{0.002\textwidth}
\end{figure}

\section{Experiments}
\label{sec:experiments}

We experiment IB-GAN on various datasets. For quantitative evaluation, we measure the disentanglement metrics proposed in \citep{Kim:2018th} on dSprites  \citep{Higgins:2016vm} and Color-dSprites
\citep{Burgess:2018uf, google1} dataset. For qualitative evaluation, we visualize latent traversal results of IB-GAN and measure FID scores~\cite{incep:43022} on CelebA \citep{liu2015faceattributes} and 3D Chairs \citep{Aubry14} dataset.

\subsubsection{Architecture.}
We follow DCGAN~\citep{Radford:2015wf} with batch normalization~\citep{batchnorm:15} for both generator and discriminator of IB-GAN.
We let the reconstructor $q_\phi(z|x)$ share the same front-end features with the discriminator $D(x)$ for the efficient use of parameters as in the conventional InfoGAN~\citep{Chen:2016tp} model.
Also, an MLP-based representation encoder $e_\psi(r|z)$ is used before the generator $G(r)$. 
Optimization is performed with RMSProp~\citep{Tieleman2012} with a momentum of 0.9. 
The batch size is 64 in all experiments. We constrain true and synthetic images to be normalized as $[-1, 1]$. Lastly, we use almost identical architecture for the generator, discriminator, reconstructor, and representation encoder in all of our experiments, except the different sizes of channel parameters depending on the datasets.
We defer more details of the IB-GAN architecture to Appendix.

\begin{figure*}[t!]
    \centering
    \includegraphics[width=\textwidth]{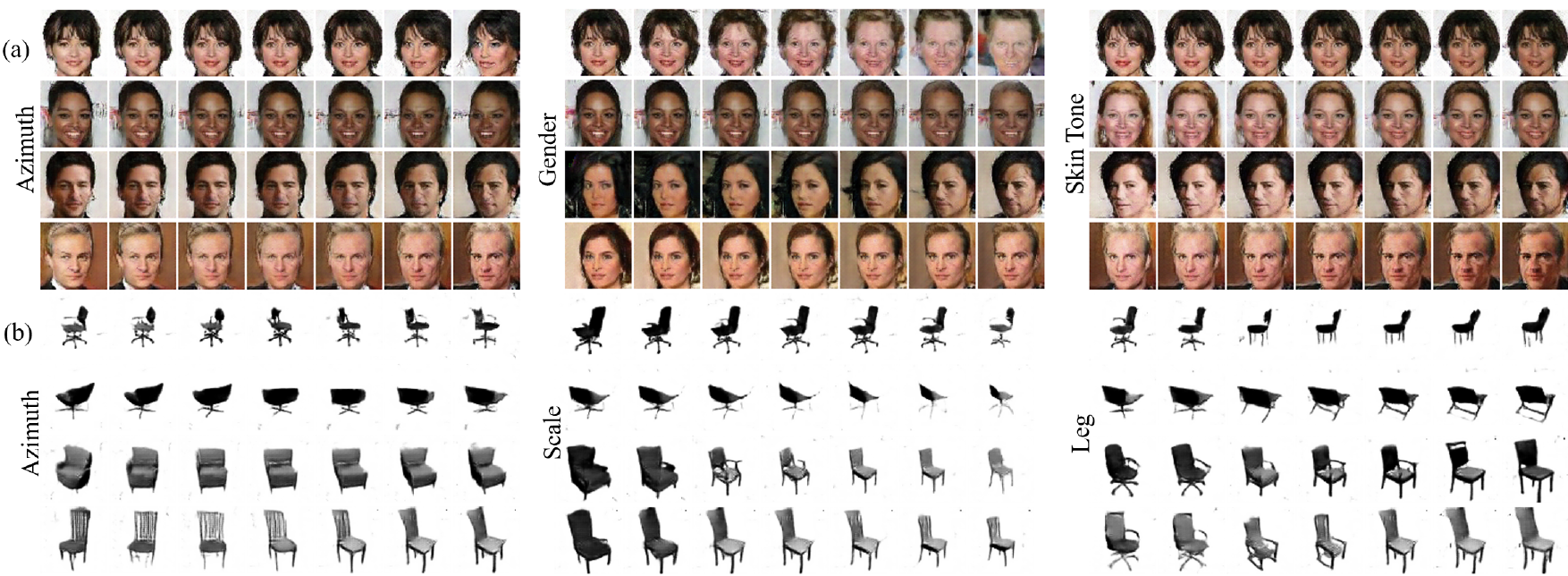}
    \caption{Latent traversals of IB-GAN that captures the factors of (a) azimuth, gender and skin tone attributes on CelebA and (b) scale, leg and azimuth on 3D Chairs. More factors captured by IB-GAN are presented in Appendix.}
        \label{fig:test2}
\end{figure*}

\begin{figure}[ht!]
    \centering
    \includegraphics[width=0.47\textwidth]{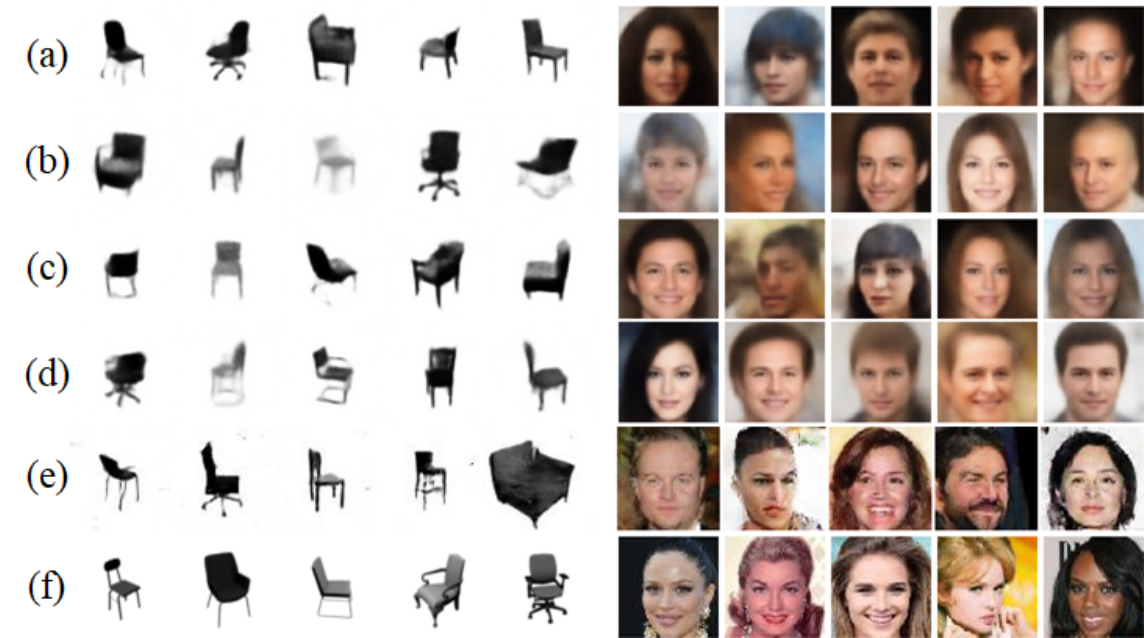}
    \caption{Comparison of random samples on CelebA and 3D Chairs dataset: (a) VAE, (b) $\beta$-VAE, (c) FactorVAE, (d) $\beta$-TCVAE, and (e) IB-GAN and (f) real images.}
    \label{fig:ramdomsalples}
    \vspace{-7pt}
\end{figure}

\subsection{Quantitative Results}
\label{sec:quantitative}

Although it is not easy to evaluate the disentanglement of representation, some quantitative metrics \citep{Higgins:2016vm,Kim:2018th, Chen:2018wu} have been proposed based on the synthetic datasets that provide ground-truth generative factors such as dSprites~\citep{Higgins:2016vm} or Color-dSprites \citep{Burgess:2018uf, google1}.
We evaluate our approach with the metric of \citep{Kim:2018th} on the dSprites and Color-dSprites datasets since many other state-of-the-art models are evaluated in this setting in \citep{google1}, including standard VAE \citep{Kingma:2013tz, Rezende:2014vm}, $\beta$-VAE \citep{Higgins:2016vm}, TC-VAE \citep{Chen:2018wu} and FactorVAE \citep{Kim:2018th}. 

\subsubsection{Disentanglement performance.} According to IB (or RD) theory~\citep{Alemi:2017va}, we can set any real values to $\beta$. For the quantitative evaluation, 
we perform hyperparameter search in the range of $\beta \in [0,1]$.
We focus on investigating the effect of $\beta \in [0,1]$ on the MI and the disentangling promoting behavior. Table \ref{dsprites-mig-table} compares the disentanglement performance metric of \citet{Kim:2018th} between methods on the dSprites and Color-dSprites~\citep{Burgess:2018uf, google1} dataset. 
The optimal average disentanglement scores $0.80$ and $0.79$ on the two datasets are obtained at $\beta=0.141$ and $\beta=0.071$, respectively. In our experiment, the disentanglement scores of IB-GAN exceed those of GAN~\citep{Goodfellow:2014td}, VAE~\citep{Kingma:2013tz, Rezende:2014vm} and InfoGAN~\citep{Chen:2016tp}, and are comparable to those of $\beta$-VAE. For the VAE baselines, we follow the model architectures and experimental settings of \citep{google1}. For the GAN baselines, we use the subset of components of IB-GAN: generator and discriminator for the vanilla GAN and additional reconstructor for the InfoGAN.

\begin{table}[t]
\centering
        \setlength{\tabcolsep}{6pt}
        \renewcommand{\arraystretch}{0.84}
        \begin{tabular}{lcc} \toprule
            {\textbf{Models}}                    & \textbf{dSprites}     &\textbf{Color-dSprites} \\ 
            \midrule
            GAN             & $0.40 \pm{0.05} $ & $0.35 \pm{0.04} $\\
            InfoGAN         & $0.61 \pm{0.03} $ & $0.55 \pm{0.08} $\\ 
            IB-GAN          & $\textbf{0.80} \pm{\textbf{0.07}}$  & $\textbf{0.79} \pm{\textbf{0.05}} $ \\ 
            \midrule
            VAE             & $0.61 \pm{0.04} $ & $0.59 \pm{0.06}$\\
            $\beta$-VAE     & $0.69 \pm{0.09} $ & $0.74 \pm{0.06}$\\
            FactorVAE       & $\textbf{0.81} \pm{\textbf{0.07}} $ & $\textbf{0.82} \pm{\textbf{0.06}}$\\ 
            $\beta$-TCVAE   & $0.79 \pm{0.06} $ & $0.80 \pm{0.07}$\\ 
            \bottomrule
          \end{tabular}
\caption{Comparison of disentanglement metric values \citep{Kim:2018th}. The average scores of IB-GAN is obtained from 10 random seeds.}
\label{dsprites-mig-table}
\label{table2}
\end{table}

\subsubsection{Traversal examples.} Figure \ref{fig:dsprite}(a) presents the visual inspection of the latent traversal \citep{Higgins:2016vm} with the learned IB-GAN model on dSprites. The IB-GAN successfully learns 5 out of 5 ground-truth factors from the dSprites, including Y and X positions, scales, rotations and shapes, which align well with the KL scores in Figure \ref{fig:convergence}. Figure \ref{fig:dsprite}(b) presents that IB-GAN captures 6 out of 6 ground-truth factors additionally including the color factor on Color-dSprites~\citep{Burgess:2018uf, google1}.

\subsubsection{Convergence.} Figure \ref{fig:convergence} shows the variations of $\mbox{KL}(e(r_i|z)||m(r_i))$ for 10-dimensional $r$ (\emph{i.e.}, $i=1,\ldots,10$) over training iterations on dSprites when $\beta=0.212$.
The increasing curves indicate that information capture by $r_i$ for $x$ increases as the learned representation is informative to reconstruct the input correctly.
Moreover, each $r_i$ increases at different points; it implies that the variational encoder $e_\psi(r|z)$ of IB-GAN slowly adapts to capture the independent factors of variations in dSprites as the upper-bound of MI increases. The similar behavior is reported in $\beta$-VAE \citep{Burgess:2018uf}. More results of convergence plots are presented in Appendix.

\subsubsection{The effect of $\beta$.} 
We inspect the effect of $\beta$ on the convergence of upper and lower MI bounds and the disentanglement score \citep{Kim:2018th} on dSprites.
We take a median value over the 150K training iterations in each trial, and then average the values over 10 different trials per $\beta$ in a range of $[0 ,1]$. Figure \ref{dsprite-beta-mig}(a) and \ref{dsprite-beta-mig}(b) illustrate the expected converged value of upper and lower MI bounds over the different $\beta$. 
When $\beta = 0$, the upper MI bound in the IB-GAN objective disappears; hence, the representation encoding $r$ can diverge from the prior distribution $m(r)$ without any restriction, resulting in a high divergence. When a small $\beta > 0$ is set, the MI upper bound constraint affects the optimization procedure. Thus the divergence between $r$ and its prior decreases drastically. After then, the MI upper bound seems to decrease gradually as the $\beta$ gets larger, consequently the lower MI bound decreases as well.
Lastly, Figure \ref{dsprite-beta-mig}(c) shows the effect of $\beta$ on the disentanglement scores. The average disentanglement score varies according to $\beta$, supporting that we could control the disentangling-promoting behavior of IB-GAN with the upper-bound of \textit{generative} MI and $\beta$.  Especially, the optimal disentanglement scores are achieved when $\beta$ is in a range of $[0.071,~0.212]$.

\subsection{Qualitative Results}
\label{sec:qualitative}

Following \citep{Chen:2016tp, Higgins:2016vm, Chen:2018wu, Kim:2018th}, we evaluate the qualitative results of IB-GAN by inspecting latent traversals. As shown in Figure \ref{fig:test2}(a), IB-GAN discovers various human recognizable attributes such as azimuth, gender, and skin tone on CelebA dataset. We also present the results of IB-GAN on 3D Chairs in Figure \ref{fig:test2}(b), where IB-GAN disentangles azimuth, scales, and leg types of chairs. These attributes are hardly captured by the original InfoGAN \citep{Chen:2016tp,Higgins:2016vm,Kim:2018th,Chen:2018wu}, demonstrating the effectiveness of the proposed model. 

Figure \ref{fig:ramdomsalples} illustrates randomly sampled images generated by IB-GAN and the VAE baselines.
Figure \ref{fig:ramdomsalples} shows that the images obtained from IB-GAN are often sharper and more realistic than those obtained from $\beta$-VAE and its variants \citep{Higgins:2016vm,Kim:2018th,Chen:2018wu}.
More qualitative results are presented in Appendix.

\begin{table}[t]
\centering

    \setlength{\tabcolsep}{1.9pt}
    \renewcommand{\arraystretch}{0.84}
        \begin{tabular}{lcc} \toprule
            {\textbf{Models}} &  CelebA & 3D Chairs \\ \midrule
            VAE             & $129.7$ & $56.2$ \\
            $\beta$-VAE     & $131.0$ & $91.3$ \\
            FactorVAE       & $109.7$ & $44.7$ \\ 
            $\beta$-TCVAE   & $125.0$ & $57.3$ \\ 
            \midrule
            GAN             & $8.4$ & $27.9 $\\
            InfoGAN         & $9.3$ & $25.6$\\ 
            IB-GAN          & $\textbf{7.4}$ & $\textbf{25.5}$ \\ \bottomrule
          \end{tabular}
\caption{FID scores on CelebA and 3D Chairs dataset. The lower FID score, the better quality, and diversity of samples.}
    \label{table:fidscore}
\end{table}

\subsubsection{FID scores.} 

In Table \ref{table:fidscore}, the FID score~\cite{incep:43022} of IB-GAN is significantly lower than those of VAEs and comparable to those of GANs, indicating that the generator of IB-GAN can produce diverse and qualitative image generation, while not only capturing various factors of variations from the dataset.
One reason for the generalization performance improvement in IB-GAN is the flexibility of learning the prior distribution. In contrast, other GAN baselines only rely on the pre-specified prior distributions. VAEs tend to degrade the reconstruction of the image due to their strong independent assumption on the representation.

\section{Conclusion} 
\label{sec:conclusion}

The proposed IB-GAN model is a new unsupervised GAN-based model for disentangled representation learning. 
Inspired by IB theory, we employ the MI minimization term to InfoGAN's objective to get the IB-GAN objective.
The resulting architecture derived from the variational inference (VI) formulation of IB-GAN's objective is partially similar to that of InfoGAN but has a critical difference; an intermediate layer of the generator is leveraged to constrain the mutual information between the input and the generated data.
The intermediate stochastic layer can serve as a learnable latent representation distribution that is trained with the generator jointly in an end-to-end fashion.
As a result, the generator of IB-GAN can harness the latent space in a disentangled and interpretable manner similar to $\beta$-VAE, while inheriting the merit of GANs (\emph{e.g.}, the model-free assumption on generators or decoders, producing good sample quality).
Our experimental results demonstrate that IB-GAN shows good performance on disentangled representation learning comparable with $\beta$-VAEs and outperforms InfoGANs. Moreover, the qualitative results also exhibit that IB-GAN can be trained to generate diverse and high-quality visual samples while capturing various factors of variations on CelebA and 3D Chairs dataset. 

\section{Acknowledgments}
This work was supported by Center for Applied Research in Artificial Intelligence(CARAI) grant funded by Defense Acquisition Program Administration(DAPA) and Agency for Defense Development(ADD) (UD190031RD). Gunhee Kim is the corresponding author. We would like to thank Byeongchang Kim and Youngjae Yu for helpful comments.

\bibliographystyle{aaai21}
\bibliography{main}

\begin{thebibliography}{58}
\providecommand{\natexlab}[1]{#1}
\providecommand{\url}[1]{\texttt{#1}}
\providecommand{\urlprefix}{URL }
\expandafter\ifx\csname urlstyle\endcsname\relax
  \providecommand{\doi}[1]{doi:\discretionary{}{}{}#1}\else
  \providecommand{\doi}{doi:\discretionary{}{}{}\begingroup
  \urlstyle{rm}\Url}\fi

\bibitem[{Achille and Soatto(2018{\natexlab{a}})}]{Achille:2017tm}
Achille, A.; and Soatto, S. 2018{\natexlab{a}}.
\newblock Emergence of invariance and disentanglement in deep representations.
\newblock \emph{JMLR} 19(1): 1947--1980.

\bibitem[{Achille and Soatto(2018{\natexlab{b}})}]{Achille:ej}
Achille, A.; and Soatto, S. 2018{\natexlab{b}}.
\newblock Information dropout: Learning optimal representations through noisy
  computation.
\newblock \emph{PAMI} 40(12): 2897--2905.

\bibitem[{Agakov and Barber(2005)}]{Barber:2006}
Agakov, F.~V.; and Barber, D. 2005.
\newblock Kernelized Infomax Clustering.
\newblock In \emph{NeurIPS}, 17--24.

\bibitem[{Alemi and Fischer(2018)}]{alemi2018gilbo}
Alemi, A.~A.; and Fischer, I. 2018.
\newblock GILBO: one metric to measure them all.
\newblock In \emph{NeurIPS}, 7037--7046.

\bibitem[{Alemi et~al.(2017)Alemi, Fischer, Dillon, and Murphy}]{Alemi:2016tba}
Alemi, A.~A.; Fischer, I.; Dillon, J.; and Murphy, K. 2017.
\newblock Deep Variational Information Bottleneck.
\newblock In \emph{ICLR}.

\bibitem[{Alemi et~al.(2018{\natexlab{a}})Alemi, Poole, Fischer, Dillon,
  Saurous, and Murphy}]{Alemi:2017va}
Alemi, A.~A.; Poole, B.; Fischer, I.; Dillon, J.; Saurous, R.~A.; and Murphy,
  K. 2018{\natexlab{a}}.
\newblock Fixing a Broken {ELBO}.
\newblock In \emph{ICML}, volume~80, 159--168.

\bibitem[{Alemi et~al.(2018{\natexlab{b}})Alemi, Poole, Fischer, Dillon,
  Saurous, and Murphy}]{Alemi:2017wn}
Alemi, A.~A.; Poole, B.; Fischer, I.; Dillon, J.; Saurous, R.~A.; and Murphy,
  K. 2018{\natexlab{b}}.
\newblock An information-theoretic analysis of deep latent-variable models.
\newblock \emph{https://openreview.net/forum?id=H1rRWl-Cb} .

\bibitem[{Aubry et~al.(2014)Aubry, Maturana, Efros, Russell, and
  Sivic}]{Aubry14}
Aubry, M.; Maturana, D.; Efros, A.~A.; Russell, B.~C.; and Sivic, J. 2014.
\newblock Seeing 3D Chairs: Exemplar Part-based 2D-3D Alignment using a Large
  Dataset of CAD Models.
\newblock In \emph{CVPR}.

\bibitem[{Barber and Agakov(2004)}]{Barber:2003uw}
Barber, D.; and Agakov, F. 2004.
\newblock The {IM} algorithm: a variational approach to information
  maximization.
\newblock In \emph{NeurIPS}, volume~16, 201.

\bibitem[{Barratt and Sharma(2018)}]{barratt2018note}
Barratt, S.; and Sharma, R. 2018.
\newblock {A Note on the Inception Score}.
\newblock \emph{arXiv preprint arXiv:1801.01973} .

\bibitem[{Bengio, Courville, and Vincent(2013)}]{bengio2013representation}
Bengio, Y.; Courville, A.; and Vincent, P. 2013.
\newblock Representation learning: A review and new perspectives.
\newblock \emph{PAMI} 35(8): 1798--1828.

\bibitem[{Burgess et~al.(2018)Burgess, Higgins, Pal, Matthey, Watters,
  Desjardins, and Lerchner}]{Burgess:2018uf}
Burgess, C.~P.; Higgins, I.; Pal, A.; Matthey, L.; Watters, N.; Desjardins, G.;
  and Lerchner, A. 2018.
\newblock Understanding disentangling in $\beta$-VAE.
\newblock \emph{arXiv preprint arXiv:1804.03599} .

\bibitem[{Chen et~al.(2018)Chen, Li, Grosse, and Duvenaud}]{Chen:2018wu}
Chen, R. T.~Q.; Li, X.; Grosse, R.~B.; and Duvenaud, D.~K. 2018.
\newblock {Isolating Sources of Disentanglement in Variational Autoencoders}.
\newblock In \emph{NeurIPS}, volume~31.

\bibitem[{Chen et~al.(2016)Chen, Duan, Houthooft, Schulman, Sutskever, and
  Abbeel}]{Chen:2016tp}
Chen, X.; Duan, Y.; Houthooft, R.; Schulman, J.; Sutskever, I.; and Abbeel, P.
  2016.
\newblock {InfoGAN: Interpretable representation learning by information
  maximizing generative adversarial nets}.
\newblock In \emph{NeurIPS}, 2180--2188.

\bibitem[{Denton and vighnesh Birodkar(2017)}]{denton2017unsupervised}
Denton, E.~L.; and vighnesh Birodkar. 2017.
\newblock {Unsupervised Learning of Disentangled Representations from Video}.
\newblock In \emph{NeurIPS}, volume~30.

\bibitem[{Donahue, Kr{\"a}henb{\"u}hl, and Darrell(2016)}]{Donahue:2016wo}
Donahue, J.; Kr{\"a}henb{\"u}hl, P.; and Darrell, T. 2016.
\newblock Adversarial feature learning.
\newblock In \emph{ICLR}.

\bibitem[{Dumoulin et~al.(2017)Dumoulin, Belghazi, Poole, Mastropietro, Lamb,
  Arjovsky, and Courville}]{Dumoulin:2016td}
Dumoulin, V.; Belghazi, I.; Poole, B.; Mastropietro, O.; Lamb, A.; Arjovsky,
  M.; and Courville, A. 2017.
\newblock Adversarially Learned Inference.
\newblock In \emph{ICLR}.

\bibitem[{Goodfellow et~al.(2014)Goodfellow, Pouget-Abadie, Mirza, Xu,
  Warde-Farley, Ozair, Courville, and Bengio}]{Goodfellow:2014td}
Goodfellow, I.; Pouget-Abadie, J.; Mirza, M.; Xu, B.; Warde-Farley, D.; Ozair,
  S.; Courville, A.; and Bengio, Y. 2014.
\newblock Generative Adversarial Nets.
\newblock In \emph{NeurIPS}, volume~27.

\bibitem[{Heusel et~al.(2017)Heusel, Ramsauer, Unterthiner, Nessler, and
  Hochreiter}]{heusel2017gans}
Heusel, M.; Ramsauer, H.; Unterthiner, T.; Nessler, B.; and Hochreiter, S.
  2017.
\newblock {GAN}s trained by a two time-scale update rule converge to a Nash
  equilibrium.
\newblock In \emph{NeurIPS}, volume~30.

\bibitem[{Higgins et~al.(2017{\natexlab{a}})Higgins, Matthey, Pal, Burgess,
  Glorot, Botvinick, Mohamed, and Lerchner}]{Higgins:2016vm}
Higgins, I.; Matthey, L.; Pal, A.; Burgess, C.; Glorot, X.; Botvinick, M.;
  Mohamed, S.; and Lerchner, A. 2017{\natexlab{a}}.
\newblock {$\beta$-VAE: Learning basic visual concepts with a constrained
  variational framework}.
\newblock In \emph{ICLR}.

\bibitem[{Higgins et~al.(2017{\natexlab{b}})Higgins, Pal, Rusu, Matthey,
  Burgess, Pritzel, Botvinick, Blundell, and Lerchner}]{Higgins:2017vt}
Higgins, I.; Pal, A.; Rusu, A.; Matthey, L.; Burgess, C.; Pritzel, A.;
  Botvinick, M.; Blundell, C.; and Lerchner, A. 2017{\natexlab{b}}.
\newblock {DARLA}: Improving Zero-Shot Transfer in Reinforcement Learning.
\newblock In \emph{ICML}, volume~70, 1480--1490.

\bibitem[{Higgins et~al.(2018)Higgins, Sonnerat, Matthey, Pal, Burgess,
  Bosnjak, Shanahan, Botvinick, Hassabis, and Lerchner}]{Higgins:2017uo}
Higgins, I.; Sonnerat, N.; Matthey, L.; Pal, A.; Burgess, C.~P.; Bosnjak, M.;
  Shanahan, M.; Botvinick, M.; Hassabis, D.; and Lerchner, A. 2018.
\newblock {SCAN:} Learning Hierarchical Compositional Visual Concepts.
\newblock In \emph{ICLR}.

\bibitem[{Hinton, Krizhevsky, and Wang(2011)}]{hinton2011transforming}
Hinton, G.~E.; Krizhevsky, A.; and Wang, S.~D. 2011.
\newblock Transforming auto-encoders.
\newblock In \emph{ICANN}.

\bibitem[{Hoffman and Johnson(2016)}]{Hoffman:vz}
Hoffman, M.~D.; and Johnson, M.~J. 2016.
\newblock {ELBO surgery: yet another way to carve up the variational evidence
  lower bound}.
\newblock In \emph{NeurIPS}, volume~1, 2.

\bibitem[{Ioffe and Szegedy(2015)}]{batchnorm:15}
Ioffe, S.; and Szegedy, C. 2015.
\newblock {Batch Normalization: Accelerating deep network training by reducing
  internal covariate shift}.
\newblock In \emph{ICML}, volume~37, 448--456.

\bibitem[{Jha et~al.(2018)Jha, Anand, Singh, and Veeravasarapu}]{Jha:2018cc}
Jha, A.~H.; Anand, S.; Singh, M.; and Veeravasarapu, V. 2018.
\newblock Disentangling factors of variation with cycle-consistent variational
  auto-encoders.
\newblock In \emph{ECCV}, 805--820.

\bibitem[{Jordan et~al.(1999)Jordan, Ghahramani, Jaakkola, and
  Saul}]{Jordan:1999kv}
Jordan, M.~I.; Ghahramani, Z.; Jaakkola, T.~S.; and Saul, L.~K. 1999.
\newblock An Introduction to Variational Methods for Graphical Models.
\newblock \emph{ML} 37(2): 183--233.

\bibitem[{Kim and Mnih(2018)}]{Kim:2018th}
Kim, H.; and Mnih, A. 2018.
\newblock Disentangling by factorising.
\newblock In \emph{ICML}, 2649--2658.

\bibitem[{Kingma et~al.(2014)Kingma, Mohamed, {Jimenez Rezende}, and
  Welling}]{kingma2014semi}
Kingma, D.~P.; Mohamed, S.; {Jimenez Rezende}, D.; and Welling, M. 2014.
\newblock Semi-Supervised Learning with Deep Generative Models.
\newblock In \emph{NeurIPS}, volume~27.

\bibitem[{Kingma and Welling(2014)}]{Kingma:2013tz}
Kingma, D.~P.; and Welling, M. 2014.
\newblock Auto-Encoding Variational Bayes.
\newblock In \emph{ICLR}.

\bibitem[{Kulkarni et~al.(2015)Kulkarni, Whitney, Kohli, and
  Tenenbaum}]{kulkarni2015deep}
Kulkarni, T.~D.; Whitney, W.~F.; Kohli, P.; and Tenenbaum, J. 2015.
\newblock {Deep convolutional inverse graphics network}.
\newblock In \emph{NeurIPS}.

\bibitem[{LeCun, Cortes, and Burges(2010)}]{lecun2010mnist}
LeCun, Y.; Cortes, C.; and Burges, C. 2010.
\newblock MNIST handwritten digit database.
\newblock \emph{ATT Labs} 2.
\newblock \urlprefix\url{http://yann.lecun.com/exdb/mnist/}.

\bibitem[{Liu et~al.(2015)Liu, Luo, Wang, and Tang}]{liu2015faceattributes}
Liu, Z.; Luo, P.; Wang, X.; and Tang, X. 2015.
\newblock Deep learning face attributes in the wild.
\newblock In \emph{ICCV}, 3730--3738.

\bibitem[{Locatello et~al.(2019)Locatello, Bauer, Lucic, Raetsch, Gelly,
  Sch{\"o}lkopf, and Bachem}]{google1}
Locatello, F.; Bauer, S.; Lucic, M.; Raetsch, G.; Gelly, S.; Sch{\"o}lkopf, B.;
  and Bachem, O. 2019.
\newblock Challenging common assumptions in the unsupervised learning of
  disentangled representations.
\newblock In \emph{ICML}, 4114--4124. PMLR.

\bibitem[{Lučić et~al.(2018)Lučić, Kurach, Michalski, Gelly, and
  Bousquet}]{are46506}
Lučić, M.; Kurach, K.; Michalski, M.; Gelly, S.; and Bousquet, O. 2018.
\newblock {Are GANs Created Equal? A Large-Scale Study}.
\newblock In \emph{NeurIPS}.

\bibitem[{Makhzani and Frey(2017)}]{Makhzani:2017ui}
Makhzani, A.; and Frey, B.~J. 2017.
\newblock {PixelGAN Autoencoders}.
\newblock In \emph{NeurIPS}, volume~30.

\bibitem[{Mathieu et~al.(2019)Mathieu, Rainforth, Siddharth, and
  Teh}]{mathieu2019disentangling}
Mathieu, E.; Rainforth, T.; Siddharth, N.; and Teh, Y.~W. 2019.
\newblock Disentangling disentanglement in variational autoencoders.
\newblock In \emph{ICML}, 4402--4412.

\bibitem[{Mathieu et~al.(2016{\natexlab{a}})Mathieu, Zhao, Zhao, Ramesh,
  Sprechmann, and LeCun}]{Mathieu:2016df}
Mathieu, M.~F.; Zhao, J.~J.; Zhao, J.; Ramesh, A.; Sprechmann, P.; and LeCun,
  Y. 2016{\natexlab{a}}.
\newblock Disentangling factors of variation in deep representation using
  adversarial training.
\newblock In \emph{NeurIPS}.

\bibitem[{Mathieu et~al.(2016{\natexlab{b}})Mathieu, Zhao, Zhao, Ramesh,
  Sprechmann, and LeCun}]{mathieu2016disentangling}
Mathieu, M.~F.; Zhao, J.~J.; Zhao, J.; Ramesh, A.; Sprechmann, P.; and LeCun,
  Y. 2016{\natexlab{b}}.
\newblock {Disentangling factors of variation in deep representation using
  adversarial training}.
\newblock In \emph{NeurIPS}, volume~29.

\bibitem[{Mescheder, Geiger, and Nowozin(2018)}]{pmlr-v80-mescheder18a}
Mescheder, L.; Geiger, A.; and Nowozin, S. 2018.
\newblock {Which Training Methods for {GAN}s do actually Converge?}
\newblock In \emph{ICML}.

\bibitem[{N et~al.(2017)N, Paige, van~de Meent, Desmaison, Goodman, Kohli,
  Wood, and Torr}]{siddharth2017learning}
N, S.; Paige, B.; van~de Meent, J.-W.; Desmaison, A.; Goodman, N.; Kohli, P.;
  Wood, F.; and Torr, P. 2017.
\newblock {Learning Disentangled Representations with Semi-Supervised Deep
  Generative Models}.
\newblock In \emph{NeurIPS}, volume~30.

\bibitem[{Peng et~al.(2019)Peng, Kanazawa, Toyer, Abbeel, and
  Levine}]{peng2018variational}
Peng, X.~B.; Kanazawa, A.; Toyer, S.; Abbeel, P.; and Levine, S. 2019.
\newblock Variational Discriminator Bottleneck: Improving Imitation Learning,
  Inverse {RL}, and {GANs} by Constraining Information Flow.
\newblock In \emph{ICLR}.

\bibitem[{Radford, Metz, and Chintala(2016)}]{Radford:2015wf}
Radford, A.; Metz, L.; and Chintala, S. 2016.
\newblock Unsupervised Representation Learning with Deep Convolutional
  Generative Adversarial Networks.
\newblock In \emph{ICLR}.

\bibitem[{Reed et~al.(2014)Reed, Sohn, Zhang, and Lee}]{reed2014learning}
Reed, S.; Sohn, K.; Zhang, Y.; and Lee, H. 2014.
\newblock Learning to Disentangle Factors of Variation with Manifold
  Interaction.
\newblock In \emph{ICML}, volume~32, 1431--1439.

\bibitem[{Rezende, Mohamed, and Wierstra(2014)}]{Rezende:2014vm}
Rezende, D.~J.; Mohamed, S.; and Wierstra, D. 2014.
\newblock Stochastic Backpropagation and Approximate Inference in Deep
  Generative Models.
\newblock In \emph{ICML}, volume~32, 1278--1286.

\bibitem[{Ridgeway(2016)}]{Ridgeway:2016wp}
Ridgeway, K. 2016.
\newblock {A Survey of Inductive Biases for Factorial Representation-Learning}.
\newblock \emph{arXiv preprint arXiv:1612.05299} .

\bibitem[{Salimans et~al.(2016)Salimans, Goodfellow, Zaremba, Cheung, Radford,
  and Chen}]{salimans2016improved}
Salimans, T.; Goodfellow, I.; Zaremba, W.; Cheung, V.; Radford, A.; and Chen,
  X. 2016.
\newblock {Improved techniques for training gans}.
\newblock In \emph{NeurIPS}.

\bibitem[{Shannon(1948)}]{shannon1948mathematical}
Shannon, C.~E. 1948.
\newblock A mathematical theory of communication.
\newblock \emph{The Bell system technical journal} 27(3): 379--423.

\bibitem[{S{\o}nderby et~al.(2017)S{\o}nderby, Caballero, Theis, Shi, and
  Husz{\'{a}}r}]{Sonderby2016a}
S{\o}nderby, C.~K.; Caballero, J.; Theis, L.; Shi, W.; and Husz{\'{a}}r, F.
  2017.
\newblock {Amortised MAP Inference for Image Super-resolution}.
\newblock In \emph{ICLR}.

\bibitem[{Springenberg(2016)}]{Springenberg:2015cat}
Springenberg, J.~T. 2016.
\newblock Unsupervised and semi-supervised learning with categorical generative
  adversarial networks.
\newblock In \emph{ICLR}.

\bibitem[{Srivastava et~al.(2017)Srivastava, Valkov, Russell, Gutmann, and
  Sutton}]{Srivastava:2017tt}
Srivastava, A.; Valkov, L.; Russell, C.; Gutmann, M.~U.; and Sutton, C. 2017.
\newblock {VEEGAN: Reducing Mode Collapse in GANs using Implicit Variational
  Learning}.
\newblock In \emph{NeurIPS}, volume~30.

\bibitem[{Szegedy et~al.(2015)Szegedy, Liu, Jia, Sermanet, Reed, Anguelov,
  Erhan, Vanhoucke, and Rabinovich}]{incep:43022}
Szegedy, C.; Liu, W.; Jia, Y.; Sermanet, P.; Reed, S.; Anguelov, D.; Erhan, D.;
  Vanhoucke, V.; and Rabinovich, A. 2015.
\newblock Going Deeper with Convolutions.
\newblock In \emph{CVPR}, 1--9.

\bibitem[{Tieleman and Hinton(2012)}]{Tieleman2012}
Tieleman, T.; and Hinton, G. 2012.
\newblock {Lecture 6.5---RmsProp: Divide the gradient by a running average of
  its recent magnitude}.
\newblock COURSERA: Neural Networks for Machine Learning.

\bibitem[{Tishby, Pereira, and Bialek(1999)}]{Tishby:2000tq}
Tishby, N.; Pereira, F.~C.; and Bialek, W. 1999.
\newblock The information bottleneck method.
\newblock In \emph{The 37th annual Allerton Conference on Communication,
  Control, and Computing}, 368--377.

\bibitem[{Tishby and Zaslavsky(2015)}]{Tishby:2015cj}
Tishby, N.; and Zaslavsky, N. 2015.
\newblock Deep learning and the information bottleneck principle.
\newblock In \emph{Information Theory Workshop (ITW)}, 1--5.

\bibitem[{Wainwright and Jordan(2008)}]{Wainwright:2008du}
Wainwright, M.~J.; and Jordan, M.~I. 2008.
\newblock \emph{Graphical Models, Exponential Families, and Variational
  Inference}.
\newblock Now Publishers Inc.

\bibitem[{Watanabe(1960)}]{Watanabe:1960cp}
Watanabe, S. 1960.
\newblock Information theoretical analysis of multivariate correlation.
\newblock \emph{IBM Journal of research and development} 4(1): 66--82.

\bibitem[{Zhao, Song, and Ermon(2018)}]{zhao2018information}
Zhao, S.; Song, J.; and Ermon, S. 2018.
\newblock The Information Autoencoding Family: A Lagrangian Perspective on
  Latent Variable Generative Models.
\newblock In \emph{UAI}.

\end{thebibliography}


\clearpage
\onecolumn

\appendix

\section*{\centering Supplementary Material for ``IB-GAN: Disentangled Representation Learning with Information Bottleneck Generative Adversarial Networks''}
\addcontentsline{toc}{section}{Supplementary Material}

\vspace{2em}

\section{A. ~~ Related Work}
\label{sec:appendix_related}

\subsection{A.1 ~~ $\beta$-VAE}
\label{sec:prelim_betavae}

$\beta$-VAE \citep{Higgins:2016vm} is one of the state-of-the-art models for unsupervised disentangled representation learning.
The key idea of $\beta$-VAE 
is to multiply the KL-divergence term of the original VAE's objective \citep{Kingma:2013tz, Rezende:2014vm} by a constant $\beta \ge 1$:
\begin{align} \label{eq:obj_betavae}
\max_{p_\theta, q_\phi} \mathcal{L}_{\text{$\beta$-VAE}}  = \mathbb{E}_{p(x)}[ \mathbb{E}_{q_\phi(z|x)} [\log p_\theta({x}|z)]] -\beta \text{KL}(q_\phi(z|x)||p(z))], 
\end{align}
where the encoder $q_\phi(z|x)$ is the variational approximation to the intractable posterior $p(z|x)$, $p(z)$ is a prior for the latent representation and $p_\theta(x|z)$ is the decoder in the VAE context. 

Recently, the connection between the $\beta$-VAE and the Information Bottleneck (IB) theory has been discovered in \citep{Alemi:2016tba, Alemi:2017va}.
That is, Eq.(\ref{eq:obj_betavae}) is equivalent to the variational formulation of IB objective\footnote{The IB objective is the Eq.(3) in the manuscript (i.e. $\mathcal{L}_{\text{IB}} = I(Z,Y) - \beta I(Z,X)$).}.
Given that computing the marginal of mutual information (MI) in the IB objective is intractable, the variational lower and upper-bound based on the \textit{representational} MI\footnote{
  The mutual information (MI) based on the encoder $q_\phi(z|x)$ is referred to the \textit{representational} MI in \citep{Alemi:2017wn} (i.e.   $I_q(Z,X)=E_{q_\phi(z|x)p(x)}[q_\phi(z|x)p(x)/q_\phi(z)p(x)]$). We distinguish it from the \textit{generative} MI based on the generator $p_\theta(x|z)$ described in the manuscript.}
is derived as:
\begin{align} \label{eq:obj_ublb}
    I_q(Z,Y) \geq \mathbb{E}_{p(y)}[ \mathbb{E}_{q_\phi(z|x)} [\log p_\theta(y|z)] + H(y), \ \ \ 
    I_q(Z,X) 
    \leq \mathbb{E}_{p(x)}[\text{KL}(q_\phi(z|x)||p(z))].
\end{align}
$p(z)$ is used to approximate $q_\phi(z)$, forming the variational upper-bound\footnote{The variational inference technique relies on the positivity of the KL divergence: $\mathbb{E}_{p(\cdot)}[\log p(\cdot)] \ge \mathbb{E}_{p(\cdot)}[\log q(\cdot)]$ for any variational (or approximating) distribution $q(\cdot)$ \citep{Jordan:1999kv, Wainwright:2008du}.}, while $p_\theta(x|z)$ approximates  $q_\phi(x|z)=q_\phi(z|x)p(x)/q_\phi(z)$, forming the variational lower-bound of the MI.
Since the target in VAE is to reconstruct data $X$ from the representation $Z$, we can set $X$ instead of the target variable $Y$ in Eq.(\ref{eq:obj_ublb}).
Consequently, the variational lower-bound of IB objective, obtained from the lower and upper-bound of the MI in Eq.(\ref{eq:obj_ublb}), corresponds to the $\beta$-VAE's objective in Eq.(\ref{eq:obj_betavae}).

\subsection{A.2 ~~ GANs} 
There have been many studies for GAN-based representation learning in a (semi-) supervised way \citep{kulkarni2015deep, reed2014learning, siddharth2017learning, mathieu2016disentangling} or an unsupervised way \citep{Springenberg:2015cat, Dumoulin:2016td, Donahue:2016wo}. InfoGAN is an unsupervised GAN model that is dedicated to the disentangled representation learning. Nevertheless, its disentanglement quality based on the Kim's metric~~\citep{Kim:2018th} has been reported less comparable to that of the $\beta$-VAE \citep{Higgins:2016vm, Kim:2018th, Chen:2018wu}. We have extended InfoGAN to IB-GAN (\textit{Information Bottleneck GAN}) by adding the upper-bound of \textit{generative} MI into the InfoGAN's objective. IB-GAN inherits several advantages of GANs (\eg the model-free assumption on generators, producing good quality of samples, and the potential use of discrete latent variables).

\newpage

\begin{figure*}[ht!]
    \centering
    \includegraphics[width=\textwidth]{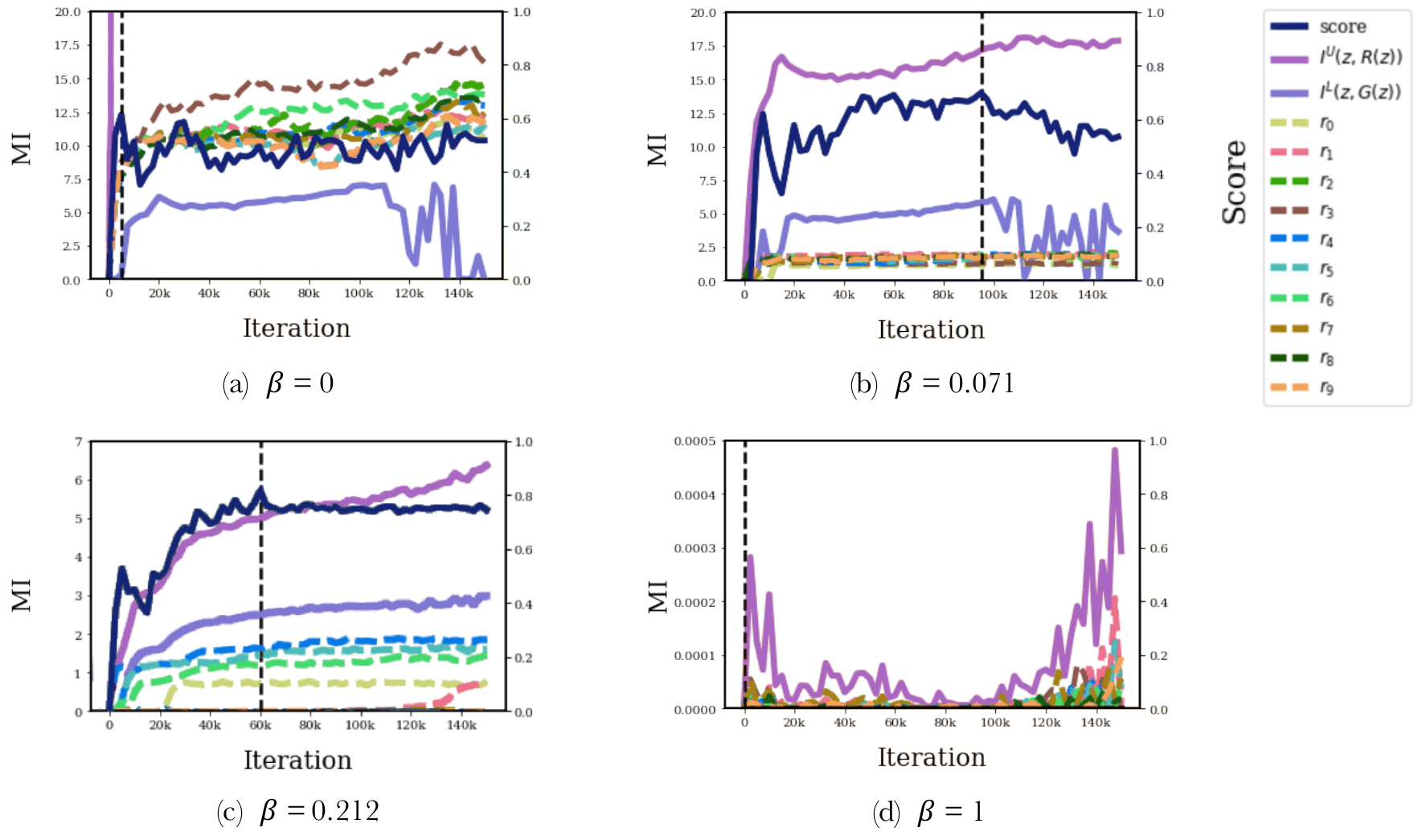}
    \caption[ ]
    {Effects of $\beta$ on the convergence of variational upper-bound and lower-bound of MI. We also depict individual KL-term $\mbox{KL}(e(r_i|z)||m(r_i))$ (dashed lines) for each $r_i$~($i=1, \cdots, 10$) over 150K training iterations. Note that the sum of all independent KL divergences is the upper-bound of MI (i.e. $I^U(z,R(z)) = \sum_{i} \mbox{KL}(e(r_i|z)||m(r_i))$). Each vertical dashed black line indicates the iteration at the highest disentanglement scores~\cite{Kim:2018th}.}
    \label{fig:effect_beta_r}
\end{figure*}

\section{B. ~~ Additional Experiments}
\label{sec:appendix_experiments}

\subsection{B.1 ~~ The effects of $\beta$ in IB-GAN on dSprites dataset}
\label{sec:appendix_beta}

To investigate the effect of $\beta$, we illustrate the convergence of the upper and lower MI bounds and the disentanglement scores in Figure \ref{fig:effect_beta_r}. The KL divergence for each independent dimension of the representation is also displayed in Figure \ref{fig:effect_beta_r}.

When $\beta=0$ as shown in Figure \ref{fig:effect_beta_r}(a), the constraining effect of the upper-bound of MI is disappeared. Hence, it is hard to distinguish the information levels captured on each independent representation. In this case, IB-GAN only has the power of maximizing the lower MI bound, similar to InfoGAN. As a result, the disentanglement score is not relatively high. 

On the other hand, when $\beta$ is 0.212, as in Figure \ref{fig:effect_beta_r}(c), both the lower and upper-bound of MI increases smoothly. The representation encoder $e_\psi(r|z)$ is slowly learned to capture the dataset's distinctive factors. Each independent KL-divergence of the representation is capped by different values. A similar behavior is observed as a key element of the disentangled representation learning in $\beta$-VAE \citep{Burgess:2018uf}. Therefore, the generator in IB-GAN with the proper $\beta$ value can learn to parsimoniously utilize each dimension of representation $r$, resulting in a good disentangled representation learning.

When $\beta=1$ as in Figure \ref{fig:effect_beta_r}(d), the upper-bound of MI drops down to almost zero, and so does the lower-bound of MI due to the constraining effect of the upper-bound. In this case, the behavior of IB-GAN is similar to the standard GAN, which yields entangled representation.

\newpage

\subsection{B.2 ~~ Latent traversal samples on dSprites dataset}
\label{sec:appendix_traversals}
\begin{figure*}[h!]
\centering\begin{tabular}{c}
\includegraphics[height=5cm, 
width=12cm]{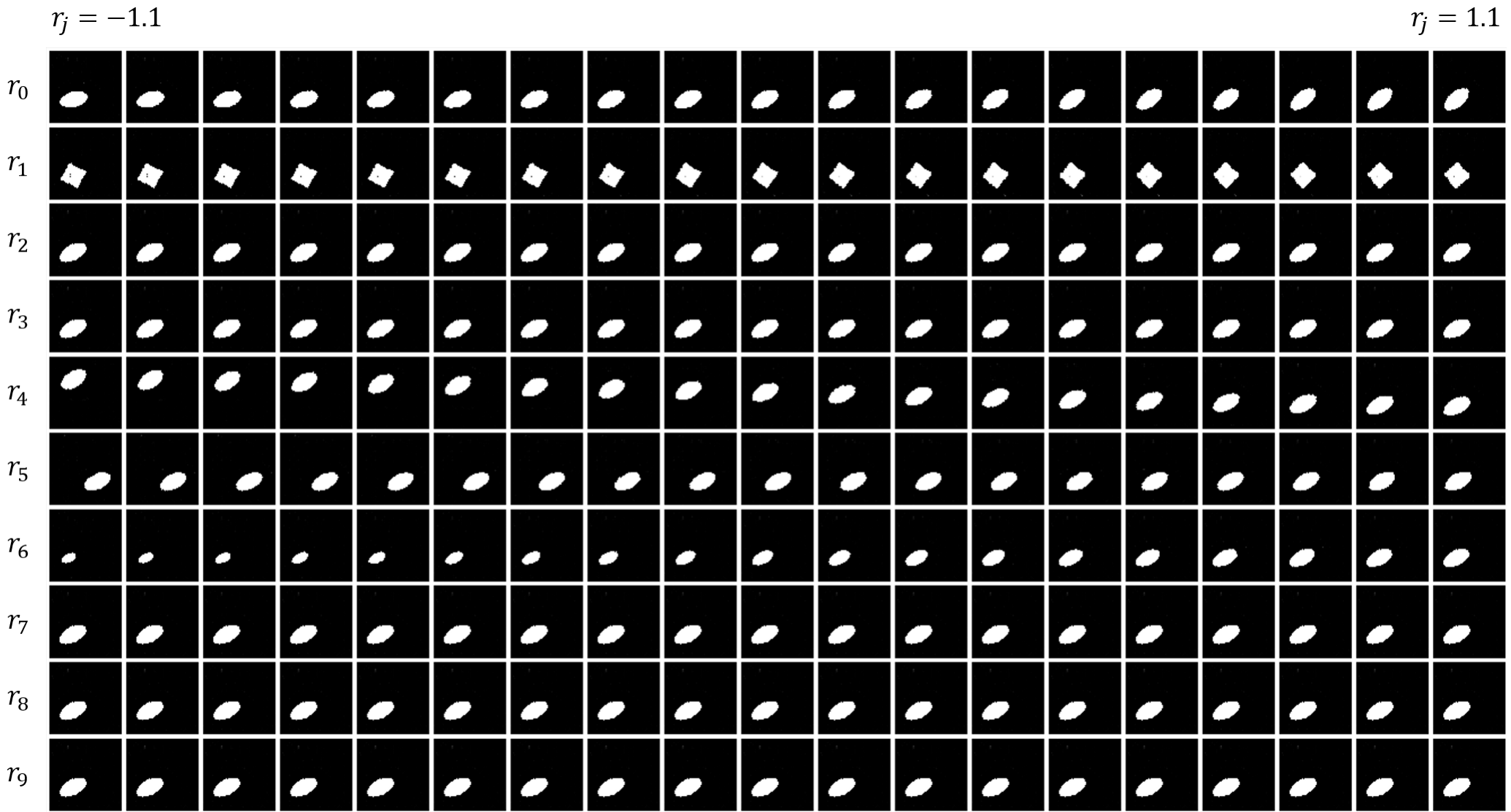}\\[0.5ex]
(a) Ellipse\\[3ex]
\vspace{-8pt}
\includegraphics[height=5cm, 
width=12cm]{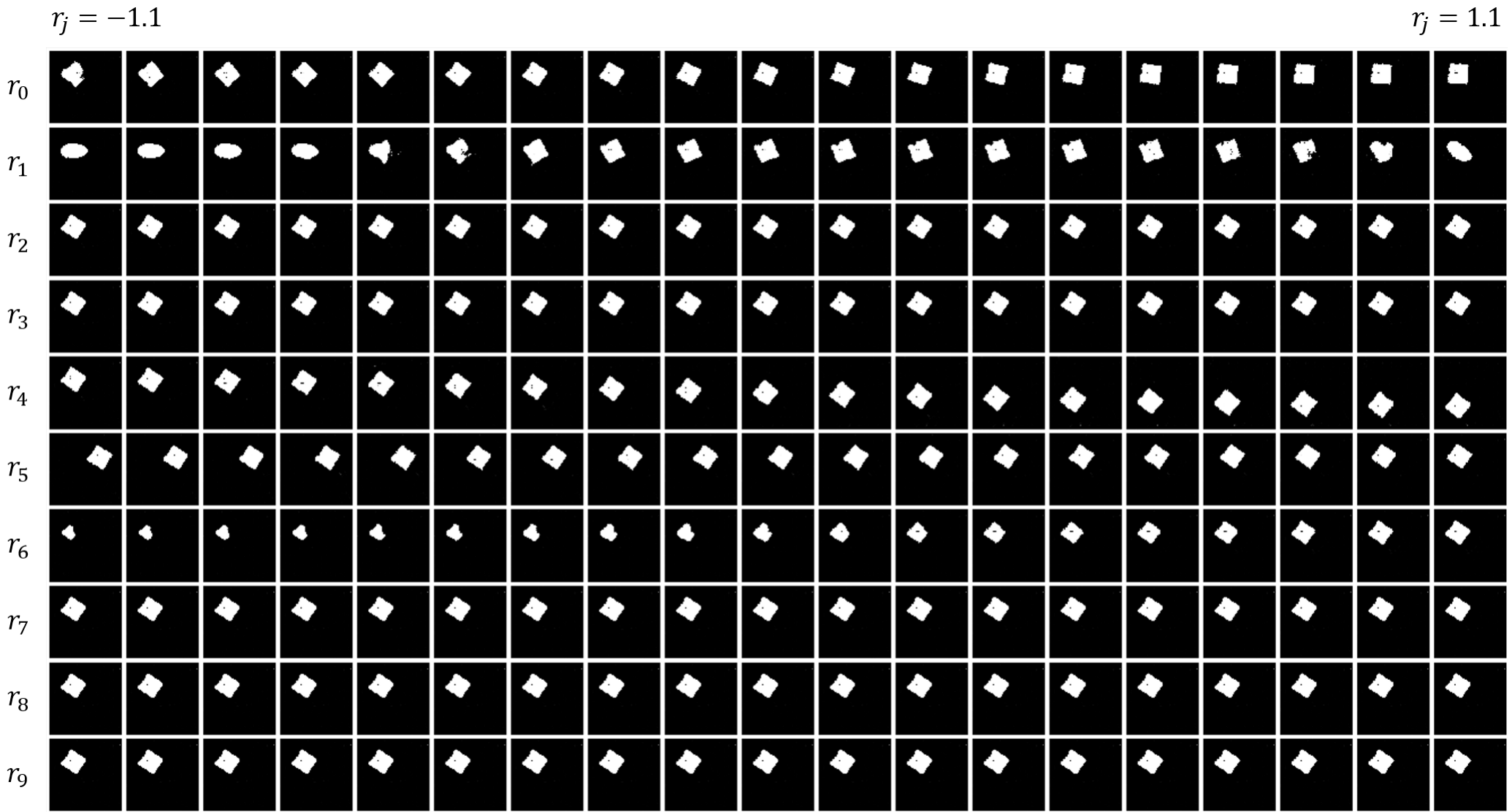}\\[1.5ex]
(b) Square\\[3ex]
\vspace{-8pt}
\includegraphics[height=5cm, 
width=12cm]{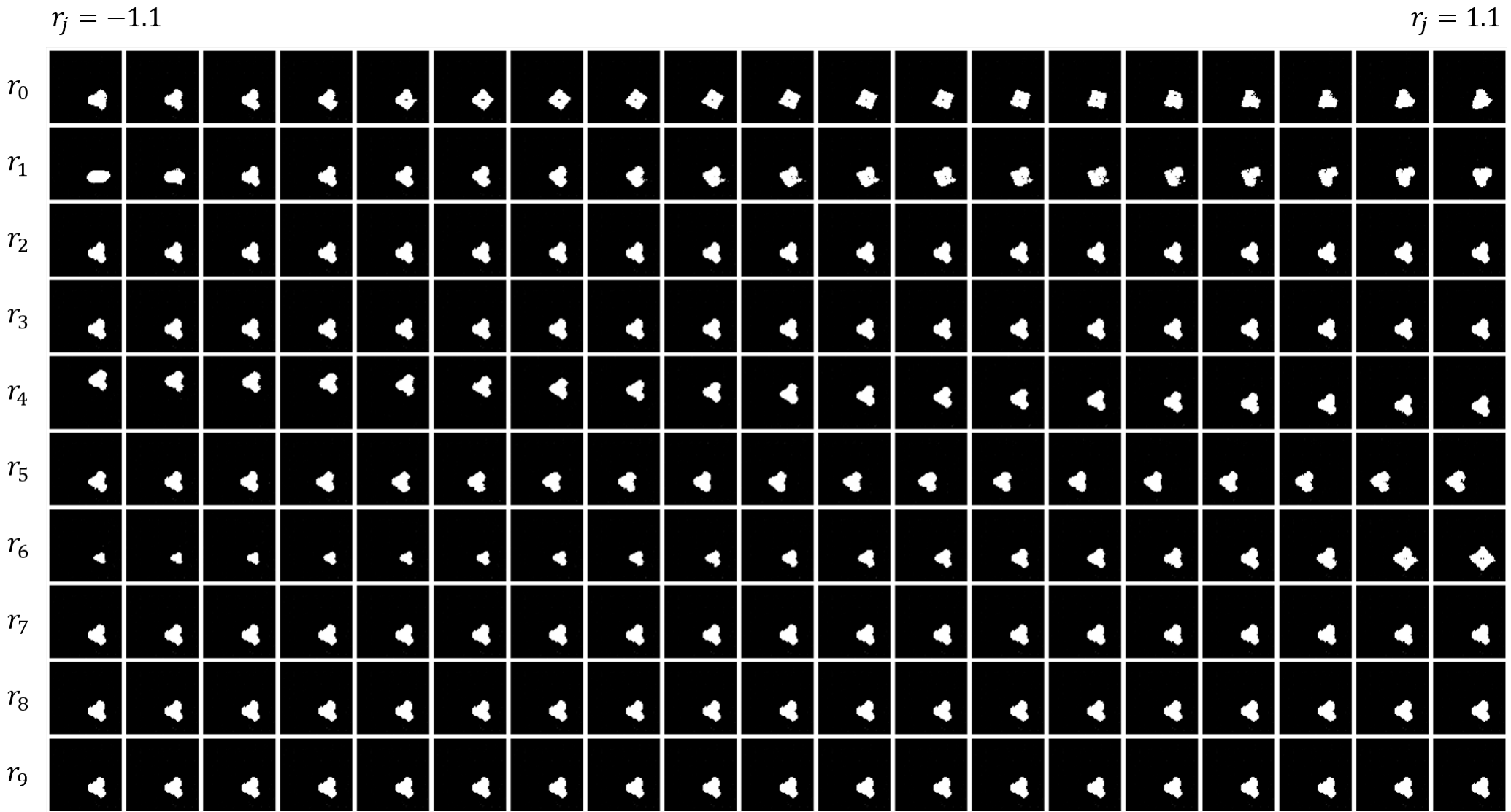}\\[1.5ex]
(c) Heart
\vspace{-8pt}
\end{tabular}
\caption{Some examples of latent traversals of three different base shapes (ellipse, square, and heart) on dSprites with the best parameter setting ($\beta=0.212$). IB-GAN successfully captures the five factors of variations: rotations~($r_0$), shapes~($r_1$), positions of $Y$~($r_4$) and $X$~($r_5$) and scales~($r_6$). The generator does not reflect the changes in $r_2, r_3, r_7, r_8 \text{ and } r_9$ since they are identical to factored zero-mean Gaussian prior $m(r_i)$ and convey no information about $z$. These results align with Figure.\ref{fig:effect_beta_r}(c); the KL-divergence values of these dimensions are nearly zero.}%
\label{fig:cont_traversal}%
\end{figure*}


\newpage
\subsection{B.3 ~~ Latent traversal samples on CelebA dataset}
\label{sec:appendix_experiments_celeba}

\begin{figure*}[ht!]
    \centering
    \begin{subfigure}[b]{0.475\textwidth}
        \centering
        \includegraphics[width=\textwidth]{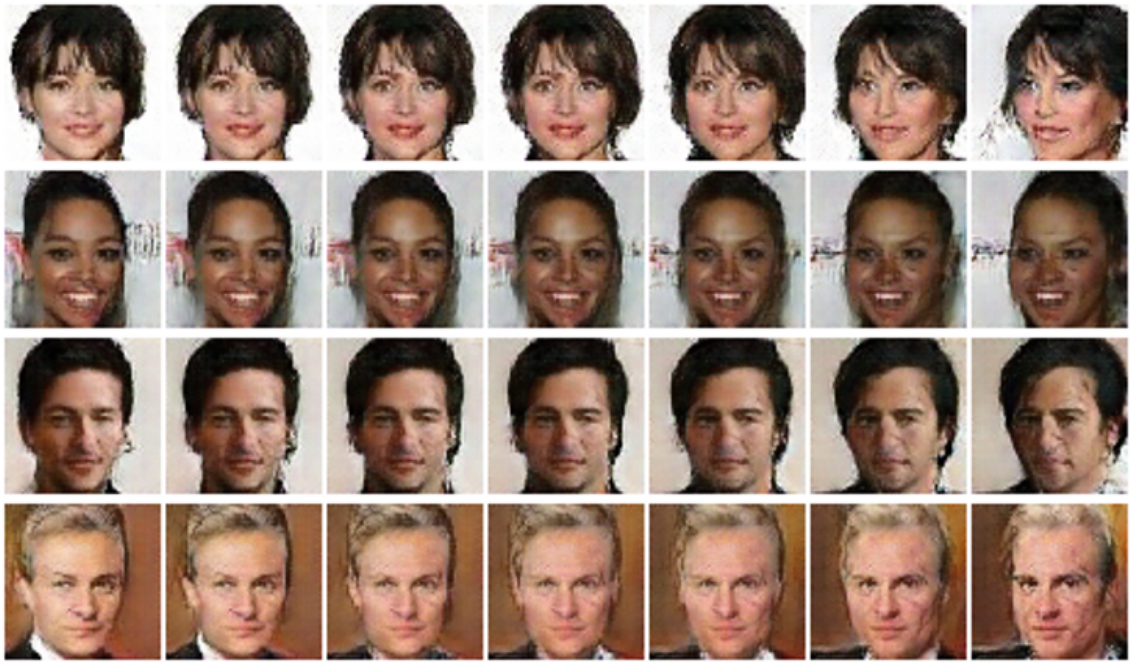}
        \caption[]%
        {{\small Azimuth}}    
        \label{fig:celeba0}
    \end{subfigure}
    \hfill
    \begin{subfigure}[b]{0.475\textwidth}  
        \centering 
        \includegraphics[width=\textwidth]{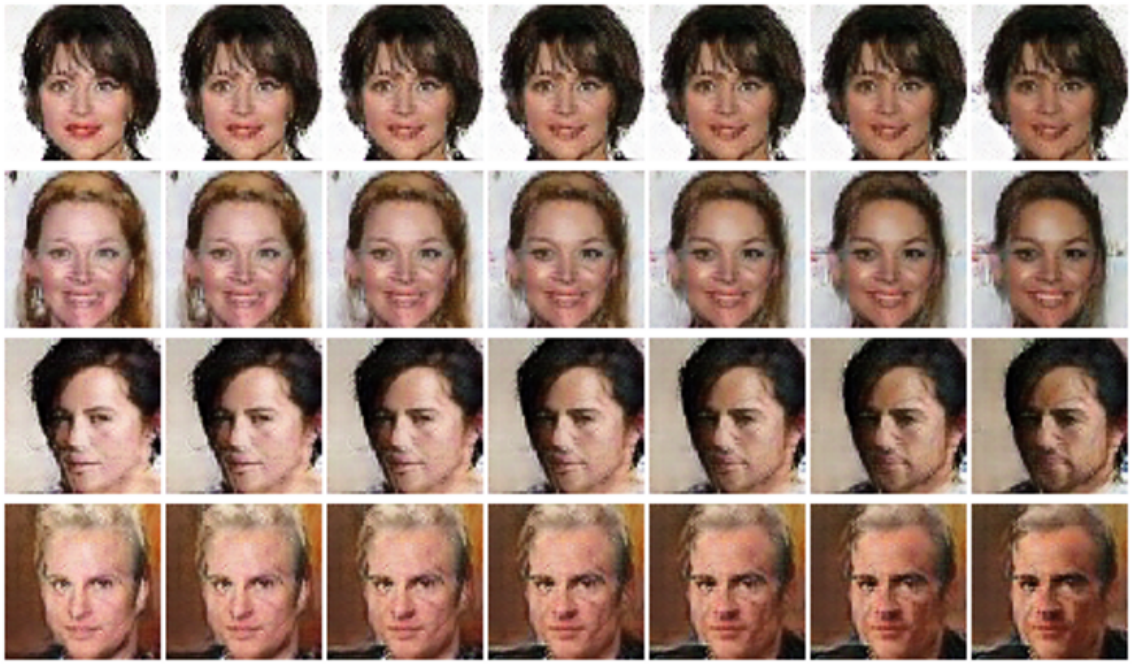} 
        \caption[]%
        {{\small Skin Tone}}    
        \label{fig:celeba1}
    \end{subfigure}
    \vskip\baselineskip
    \begin{subfigure}[b]{0.475\textwidth}   
        \centering 
        \includegraphics[width=\textwidth]{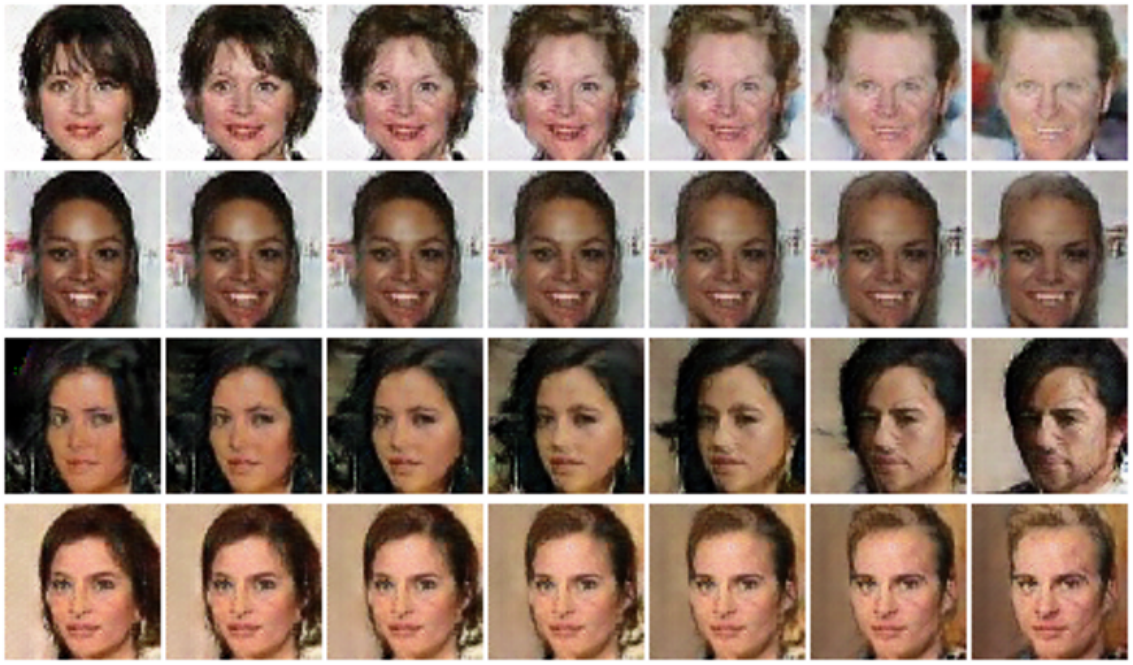} 
        \caption[]%
        {{\small Gender}}    
        \label{fig:celeba2}
    \end{subfigure}
    \hfill
    \begin{subfigure}[b]{0.475\textwidth}   
        \centering 
        \includegraphics[width=\textwidth]{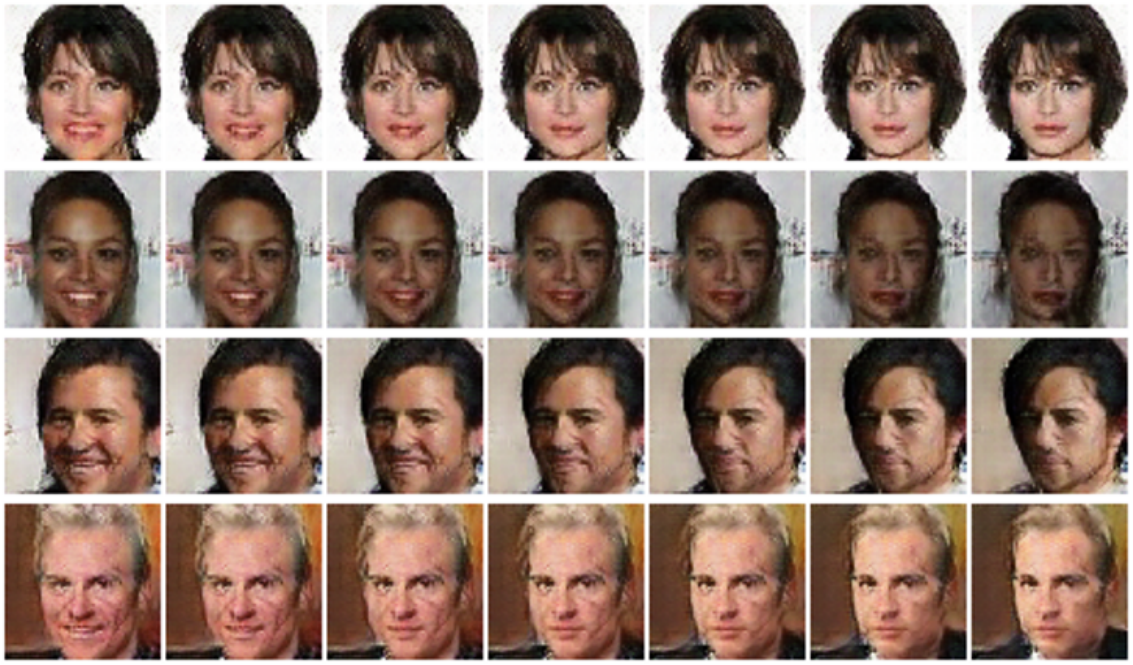} 
        \caption[]%
        {{\small Smile}}
        \label{fig:celeba3}
    \end{subfigure}
    \vskip\baselineskip
    \begin{subfigure}[b]{0.475\textwidth}   
        \centering 
        \includegraphics[width=\textwidth]{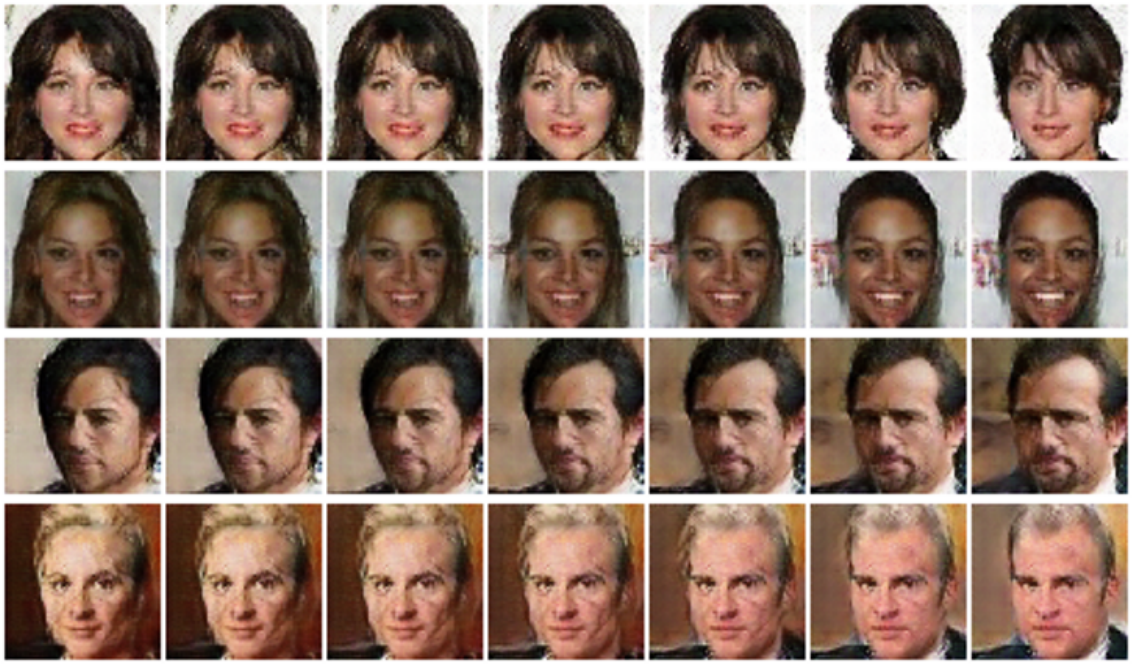} 
        \caption[]%
        {{\small Hair Length}}    
        \label{fig:celeba4}
    \end{subfigure}
    \hfill
    \begin{subfigure}[b]{0.475\textwidth}   
        \centering 
        \includegraphics[width=\textwidth]{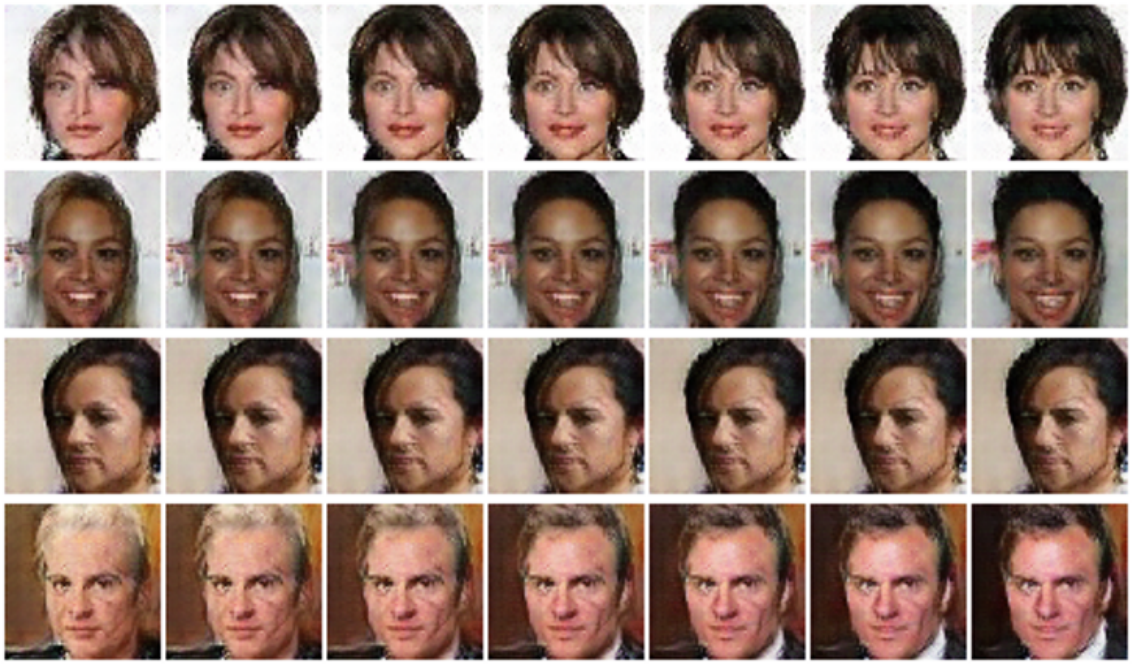} 
        \caption[]%
        {{\small Hair Color}}
        \label{fig:celeba5}
    \end{subfigure}
    \caption[ ]
    {Latent traversals of attributes captured by six different $r$ vectors on CelebA dataset with the parameter setting of $\beta=0.325$, $\gamma=2$.}
    \label{fig:celeba_traverse}
\end{figure*}

We illustrate more qualitative results of IB-GAN trained on CelebA datasets. As shown in Figure \ref{fig:celeba_traverse}, IB-GAN discovers various attributes of human faces: (a) azimuth, (b) skin tone, (c) gender, (d) smile, (e) hair length and (f) hair color. All features in Figure \ref{fig:celeba_traverse} are captured by the model trained with the parameter setting of $\beta=0.325, \gamma=2$. Note that these attributes are hardly captured in the original InfoGAN \citep{Chen:2016tp,Higgins:2016vm,Kim:2018th,Chen:2018wu}, demonstrating the effectiveness of disentangled representation learning by IB-GAN.

\subsection{B.4 ~~ Latent traversal samples on 3D Chair dataset}

\begin{figure*}[ht!]
    \centering
    \begin{subfigure}[b]{0.475\textwidth}
        \centering
        \includegraphics[width=\textwidth]{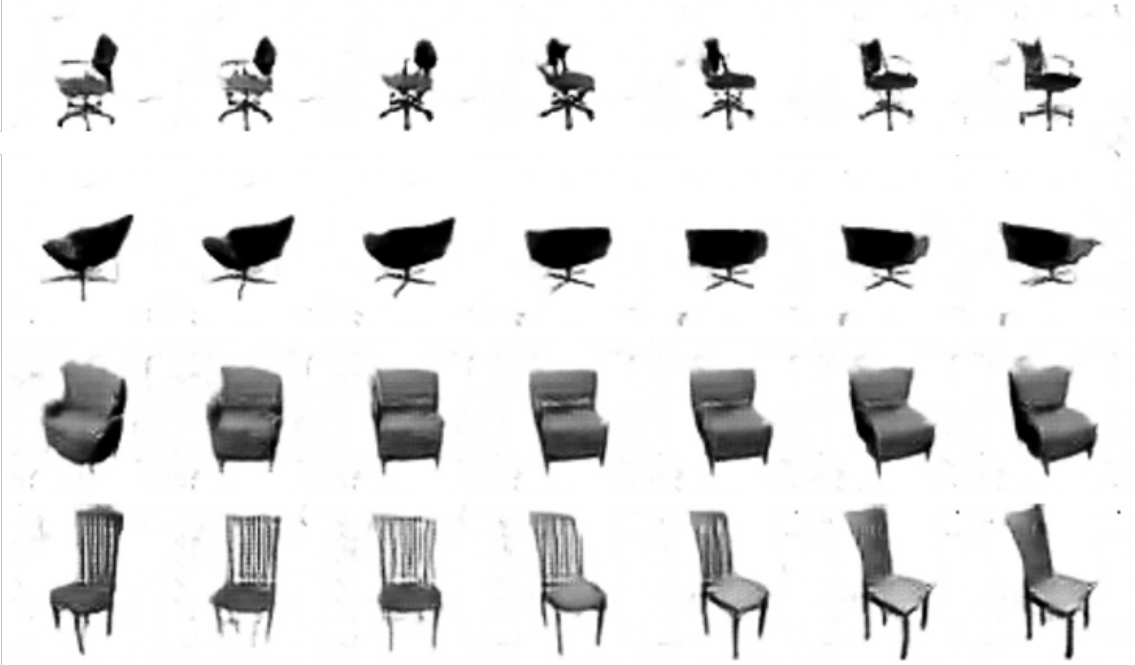}
        \caption[]%
        {{\small Azimuth}}    
        \label{fig:3dchairs0}
    \end{subfigure}
    \hfill
    \begin{subfigure}[b]{0.475\textwidth}  
        \centering 
        \includegraphics[width=\textwidth]{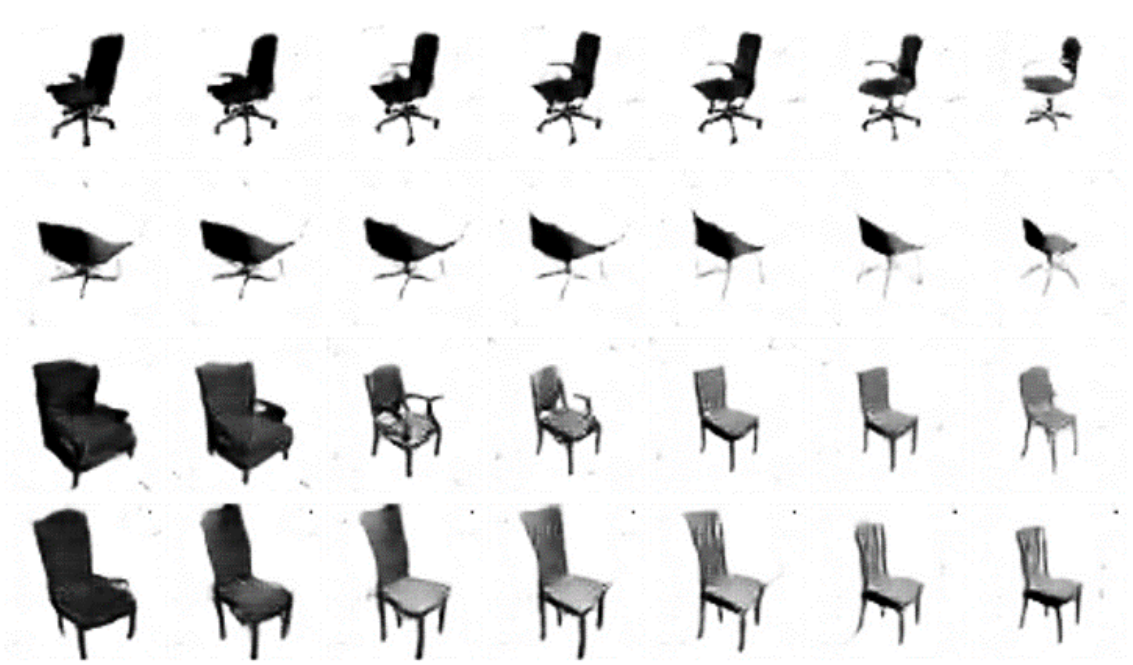} 
        \caption[]%
        {{\small Scale}}    
        \label{fig:3dchairs1}
    \end{subfigure}
    \vskip\baselineskip
    \begin{subfigure}[b]{0.475\textwidth}   
        \centering 
        \includegraphics[width=\textwidth]{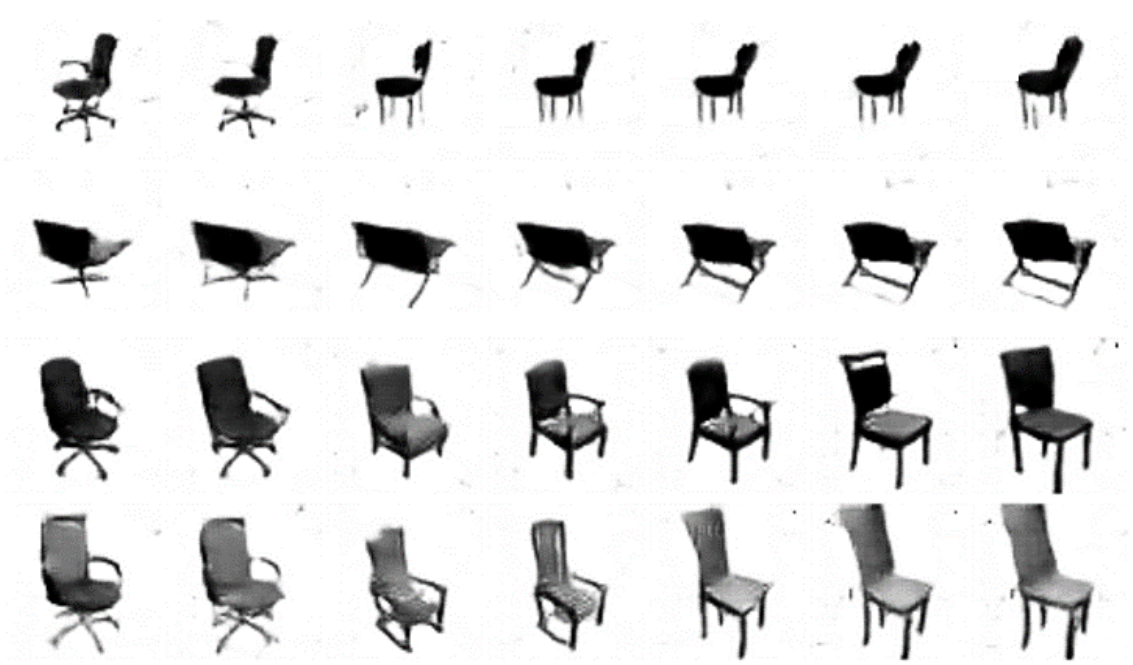} 
        \caption[]%
        {{\small Leg}}    
        \label{fig:3dchairs2}
    \end{subfigure}
    \hfill
    \begin{subfigure}[b]{0.475\textwidth}   
        \centering 
        \includegraphics[width=\textwidth]{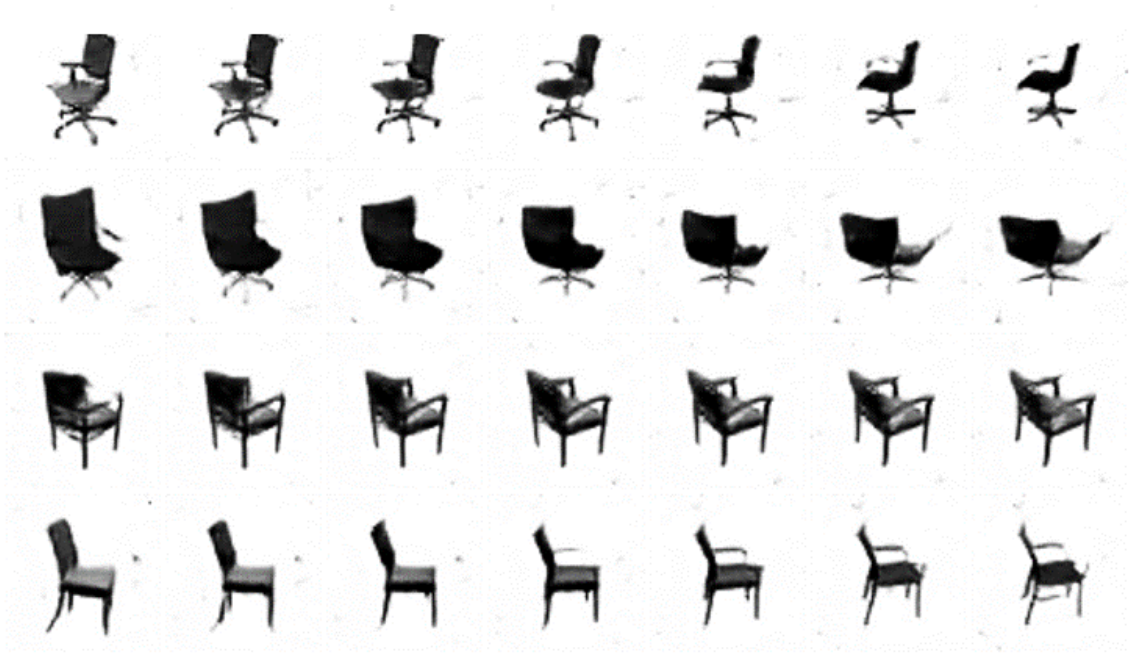} 
        \caption[]%
        {{\small Back Length}}
        \label{fig:3dchairs3}
    \end{subfigure}
    \vskip\baselineskip
    \begin{subfigure}[b]{0.475\textwidth}   
        \centering 
        \includegraphics[width=\textwidth]{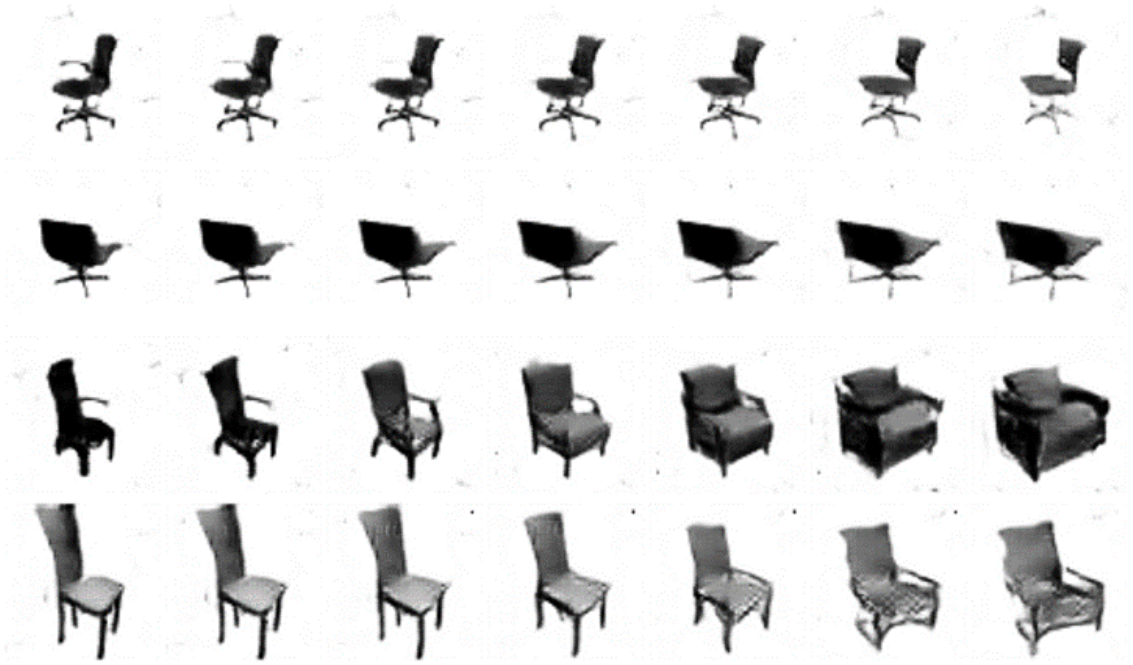} 
        \caption[]%
        {{\small Width}}    
        \label{fig:3dchairs4}
    \end{subfigure}
    \caption[ ]
    {Latent traversals of attributes captured by five different $r$ vectors on 3D Chairs dataset with the parameter setting of $\beta=0.35$, $\gamma=1.2$.}
    \label{fig:3dchairs_traverse}
\end{figure*}

For 3D Chairs dataset, 
Figure \ref{fig:3dchairs_traverse} shows that IB-GAN can capture following factors of chairs: (a) azimuth, (b) scale, (c) leg, (d) back length and (e) width. We obtain the results with the parameter setting of $\beta=0.35$ and $\gamma=1.2$.

\clearpage
\subsection{B.5 ~~ Random samples generated by the models on CelebA and 3D Chairs dataset}

\begin{figure*}[h!]
\centering
\begin{tabular}{cccc}
\includegraphics[width=0.3\textwidth]{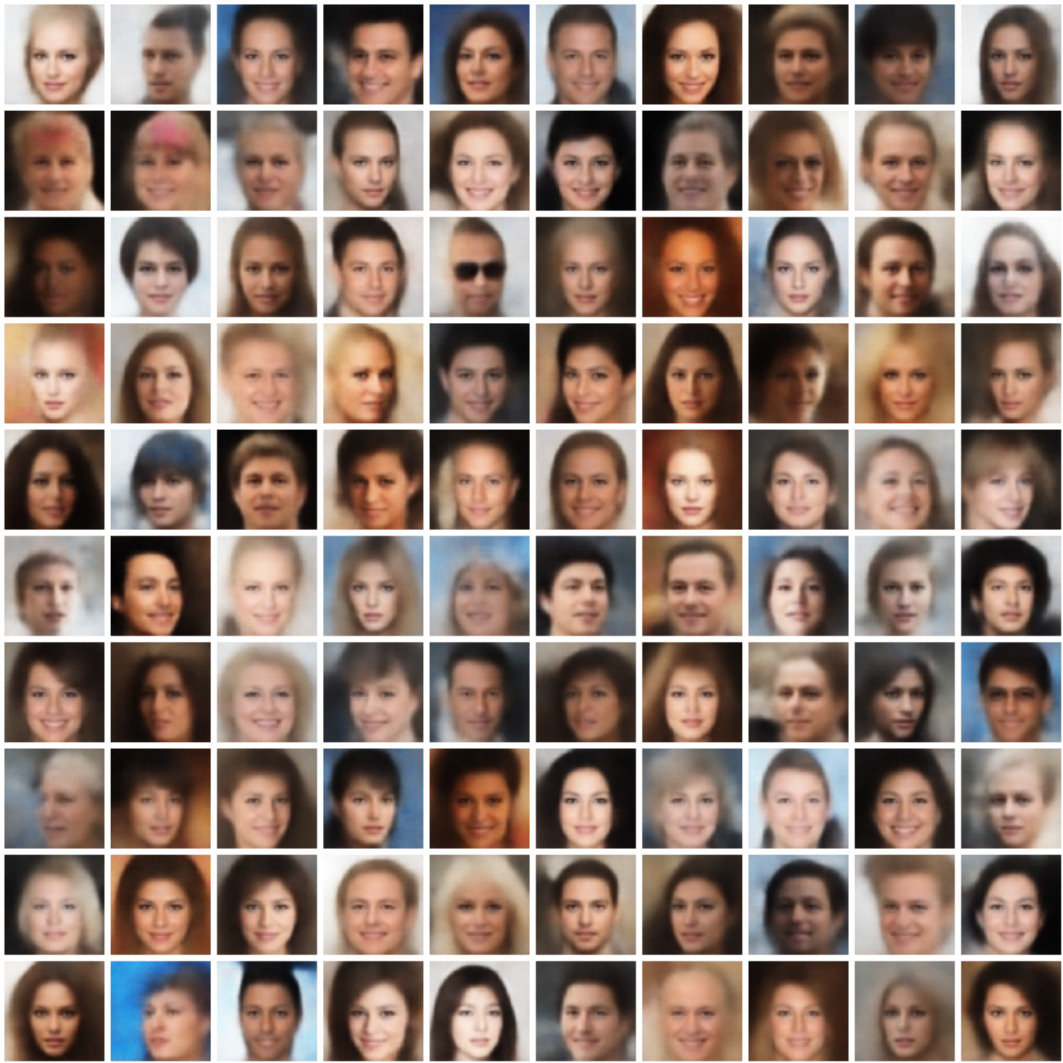} &
\includegraphics[width=0.3\textwidth]{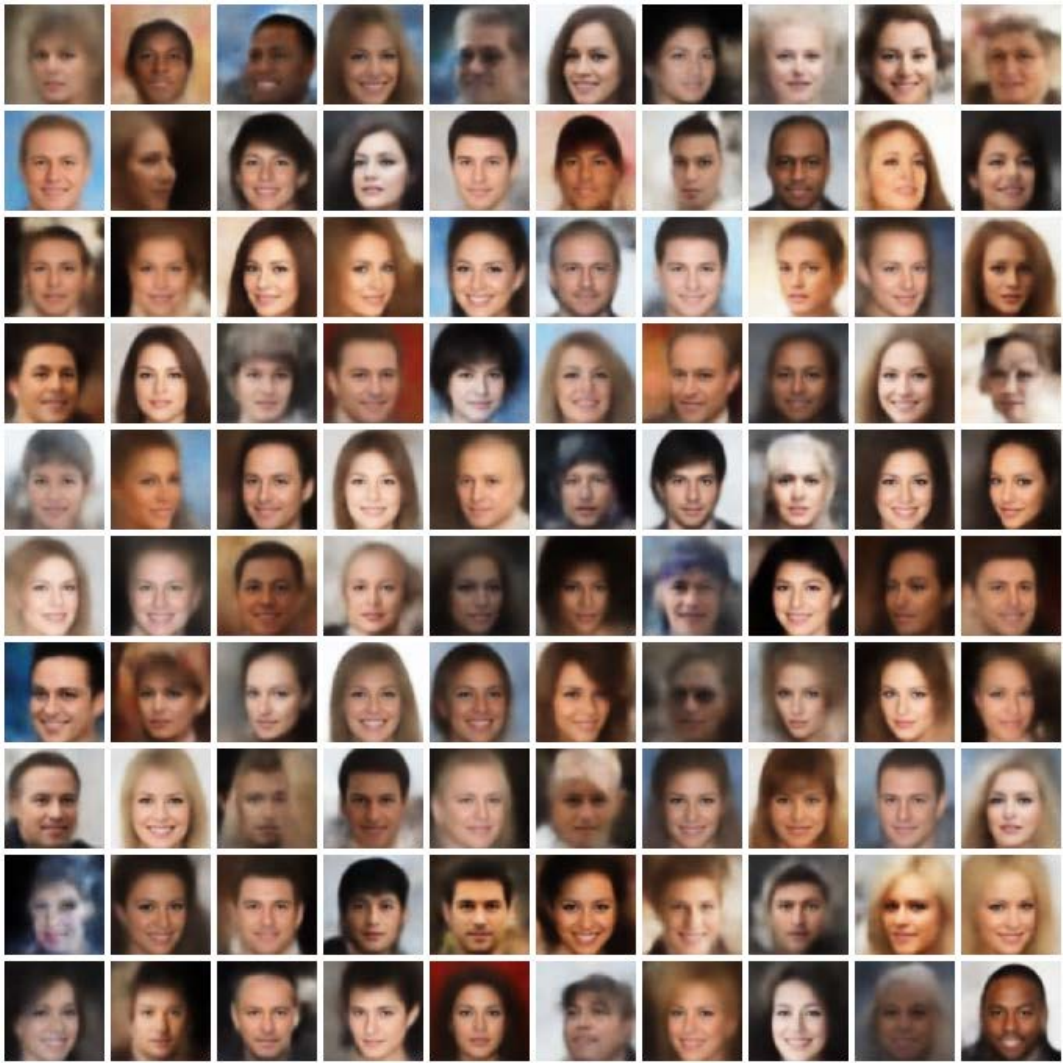} &
\includegraphics[width=0.3\textwidth]{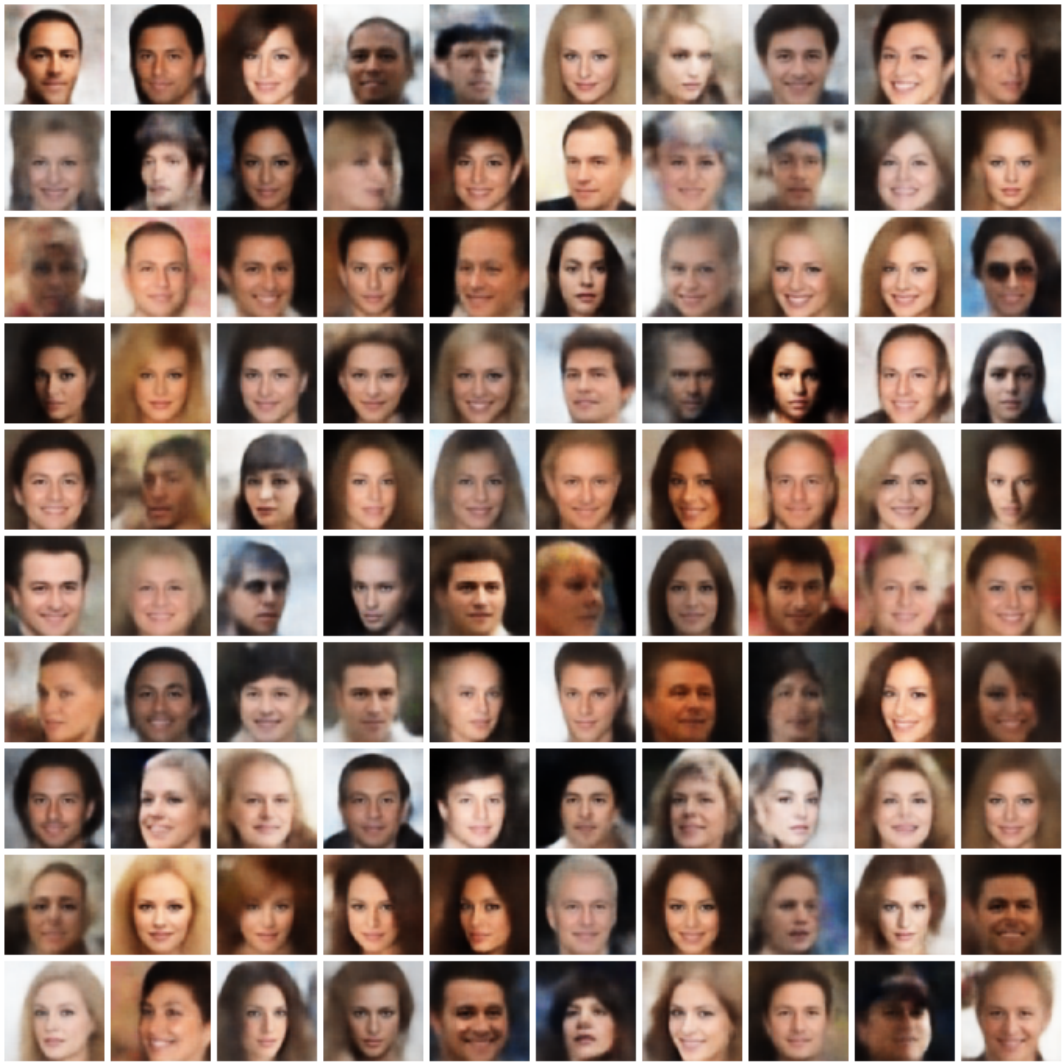} \\
(a) VAE  & (b) $\beta$-VAE & (c) FactorVAE  \\[6pt]
\end{tabular}
\begin{tabular}{cccc}
\includegraphics[width=0.3\textwidth]{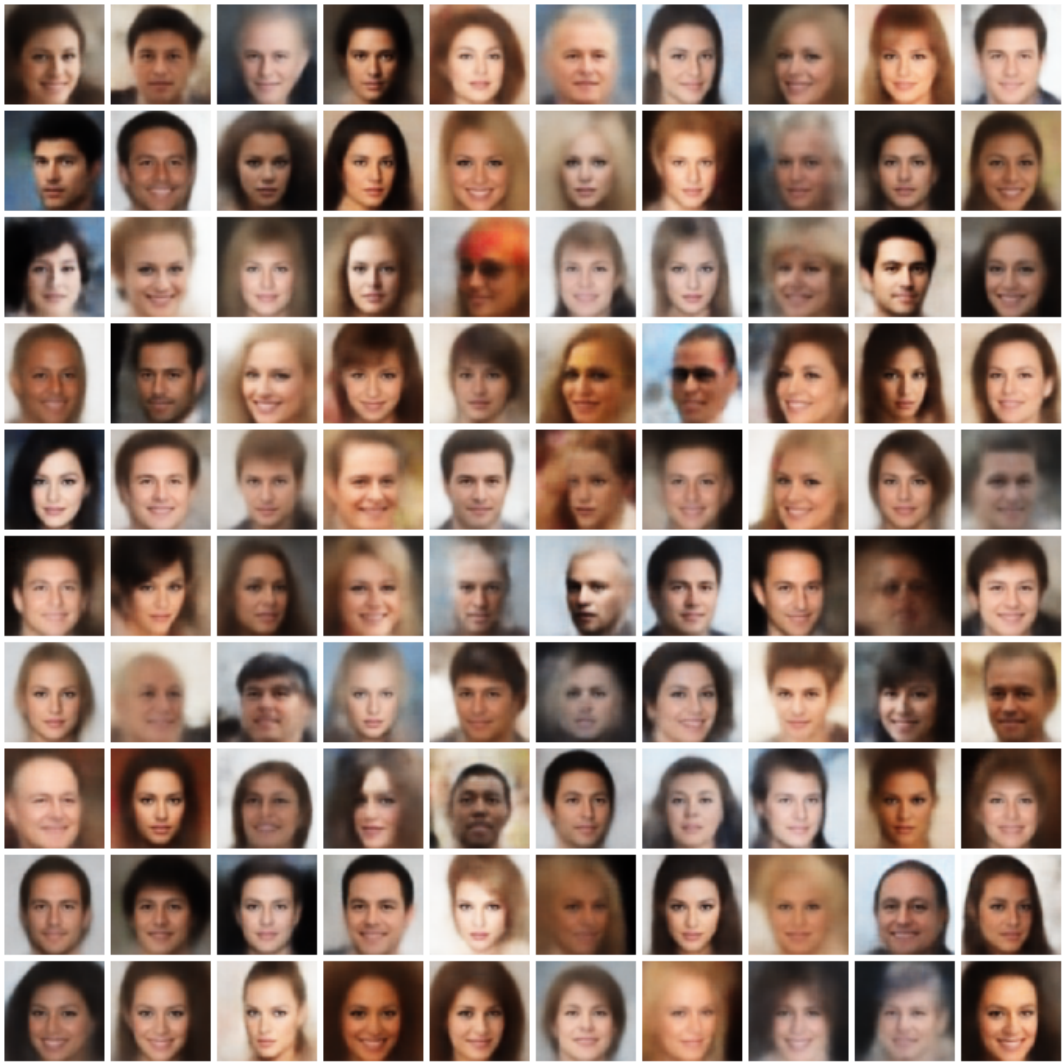} &
\includegraphics[width=0.3\textwidth]{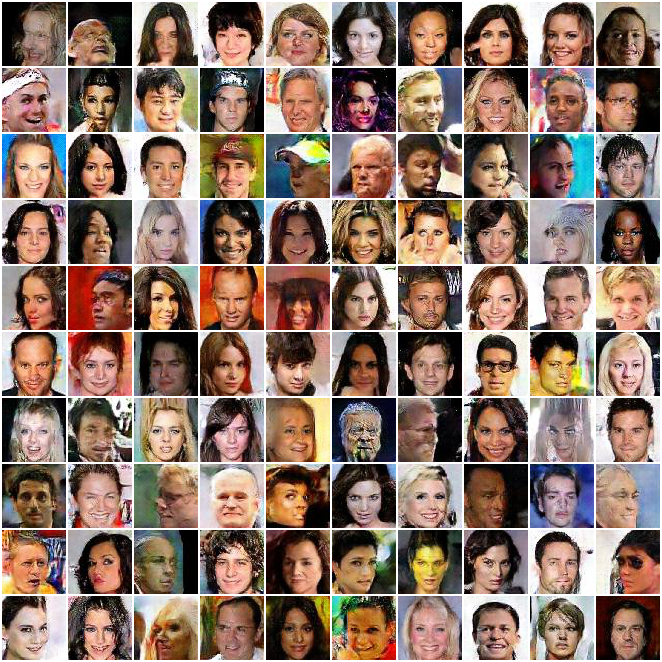} & 
\includegraphics[width=0.3\textwidth]{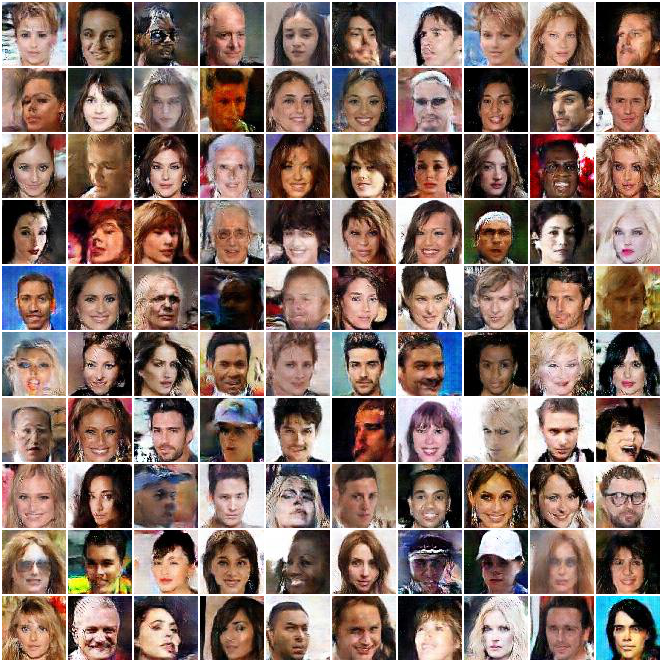} \\
(d) $\beta$-TCVAE  & (e) GAN & (f) InfoGAN\\[6pt]
\end{tabular}
\begin{tabular}{ccc}
\includegraphics[width=0.3\textwidth]{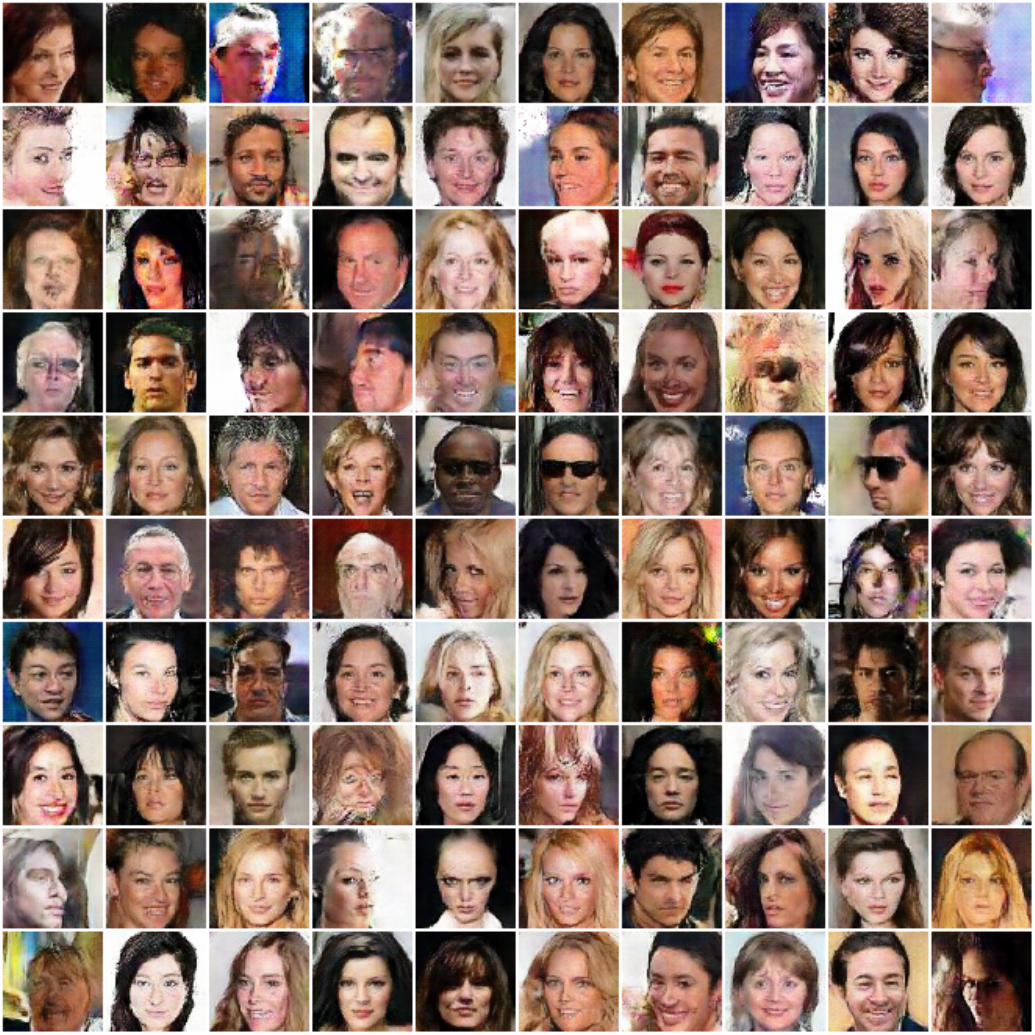} &
\includegraphics[width=0.3\textwidth]{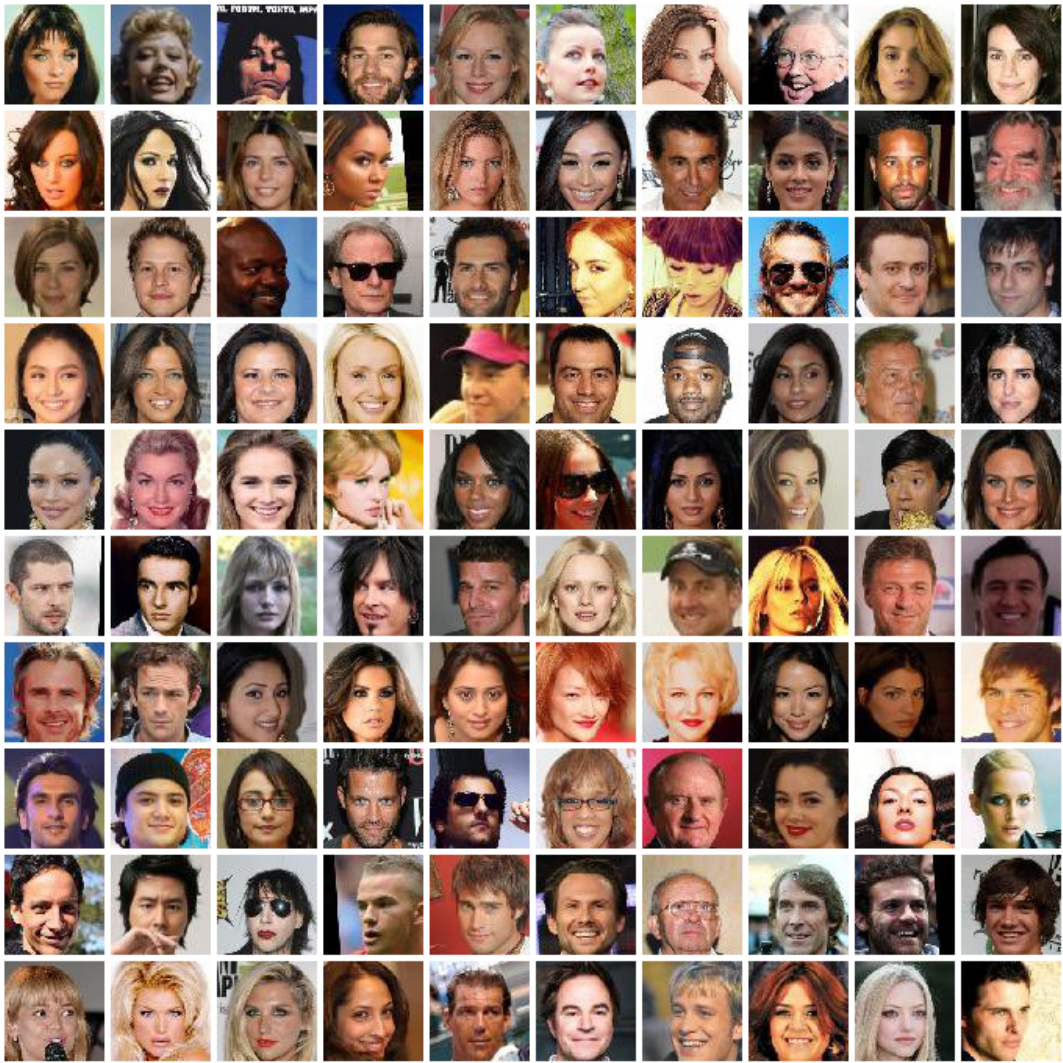} \\
(g) IB-GAN  & (h) Real Samples\\[6pt]
\end{tabular}
\caption{Comparison of randomly sampled images that are generated by the $\beta$-VAE and GAN baselines and IB-GAN on CelebA dataset.}\label{fig:celeba_comp}
\end{figure*}

\begin{figure*}[ht!]
\centering
\begin{tabular}{cccc}
\includegraphics[width=0.3\textwidth]{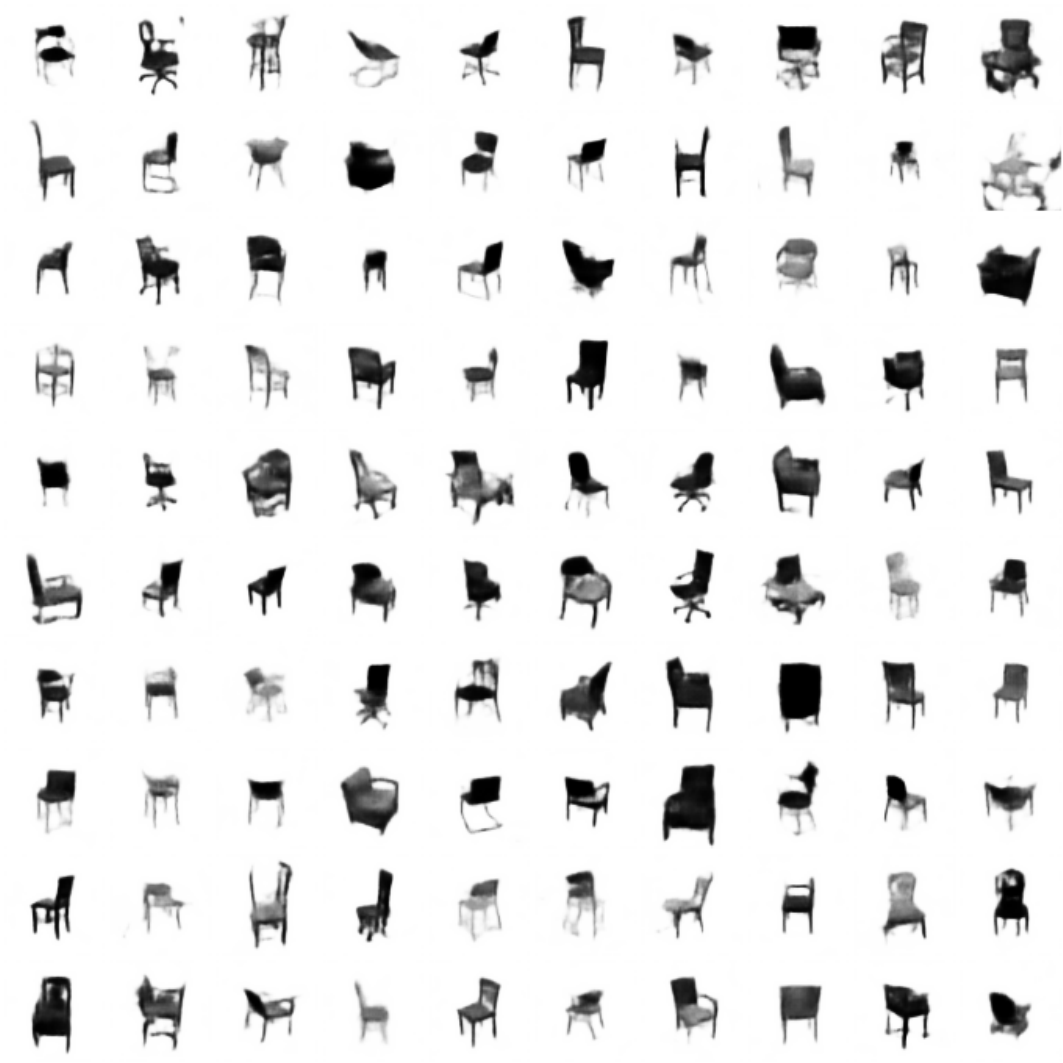} &
\includegraphics[width=0.3\textwidth]{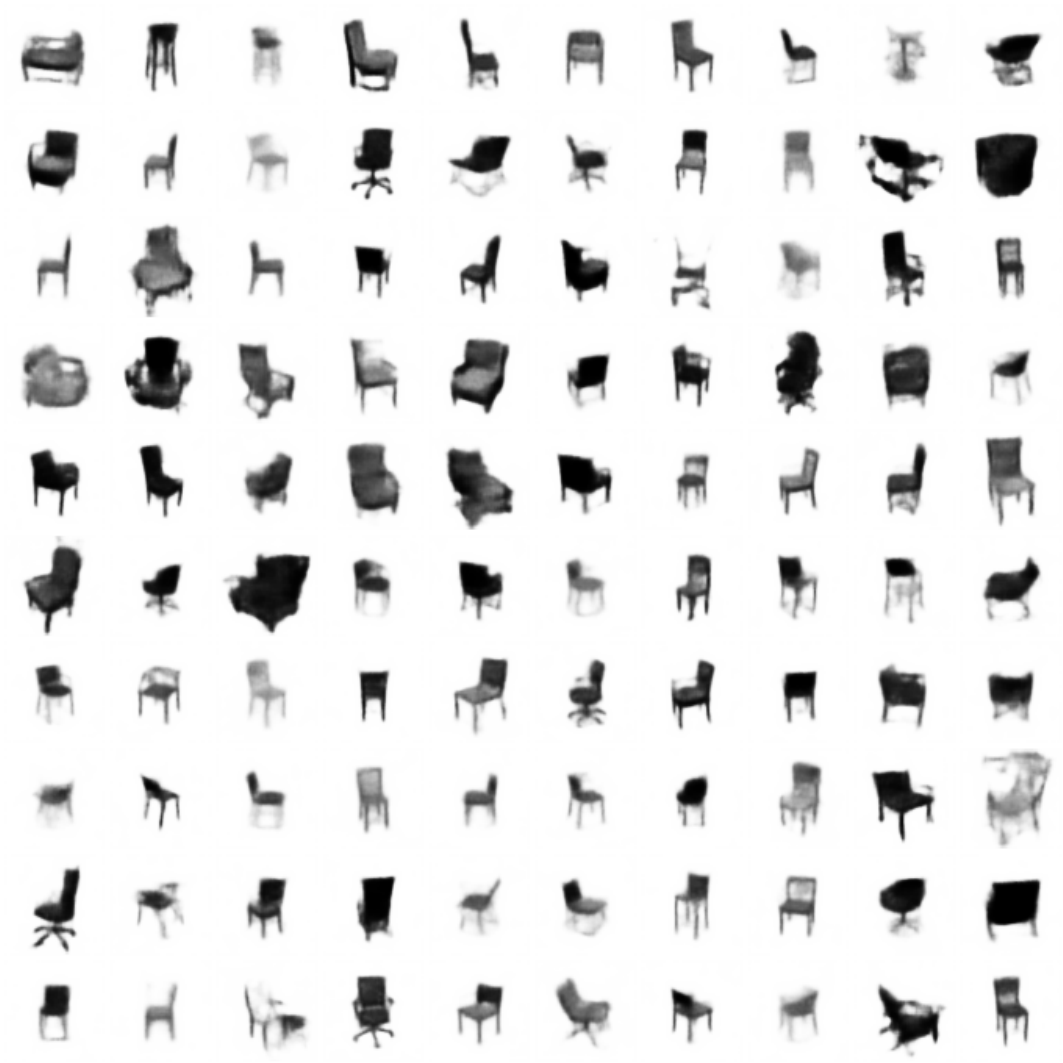} &
\includegraphics[width=0.3\textwidth]{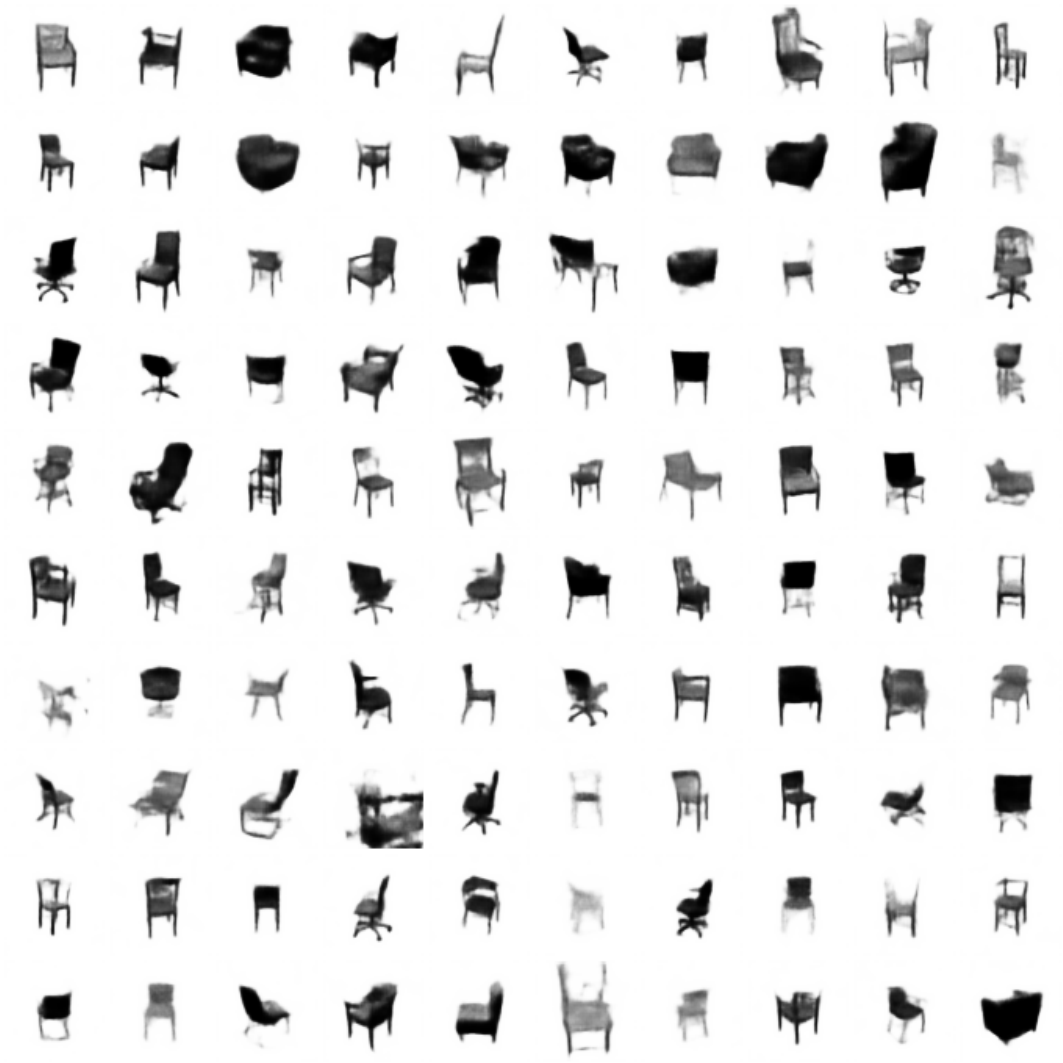} \\
(a) VAE  & (b) $\beta$-VAE & (c) FactorVAE  \\[6pt]
\end{tabular}
\begin{tabular}{cccc}
\includegraphics[width=0.3\textwidth]{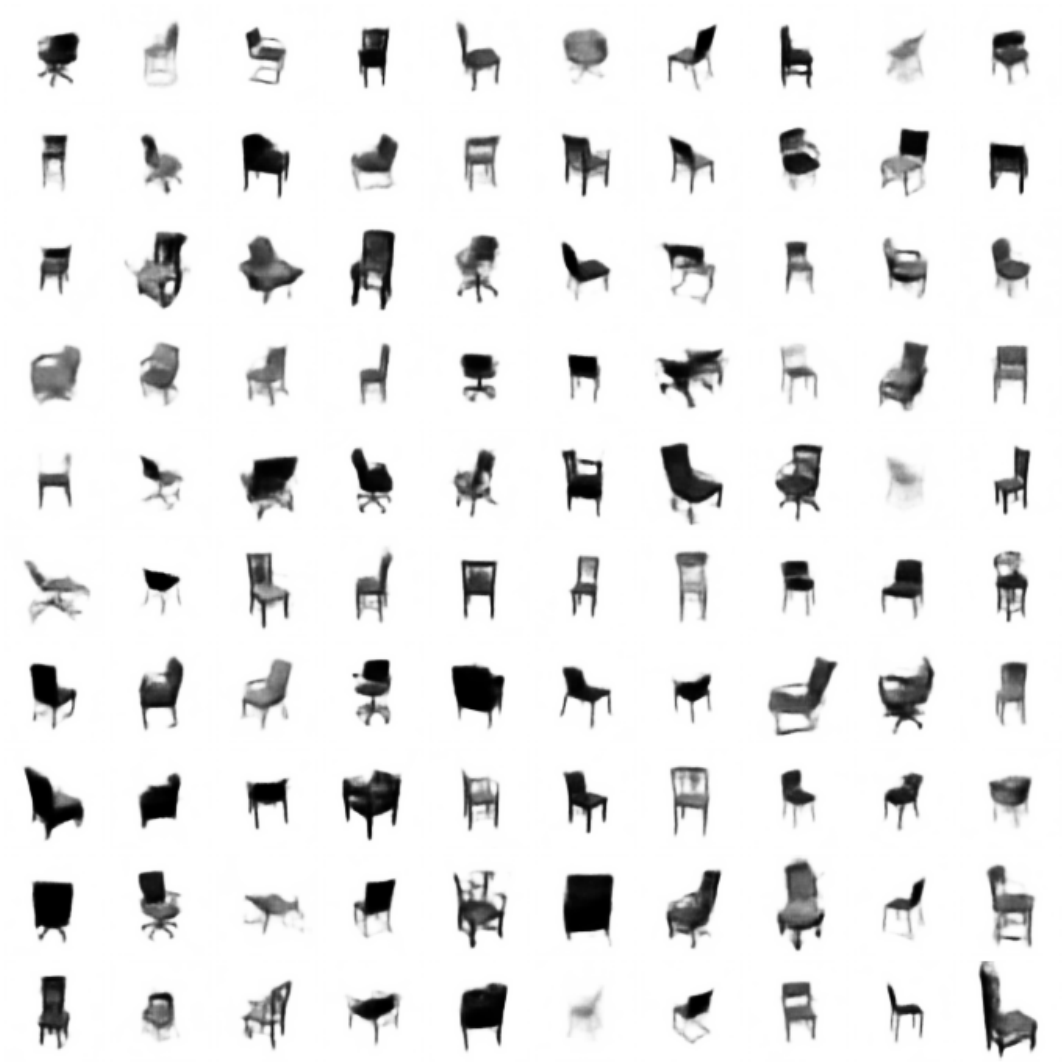} &
\includegraphics[width=0.3\textwidth]{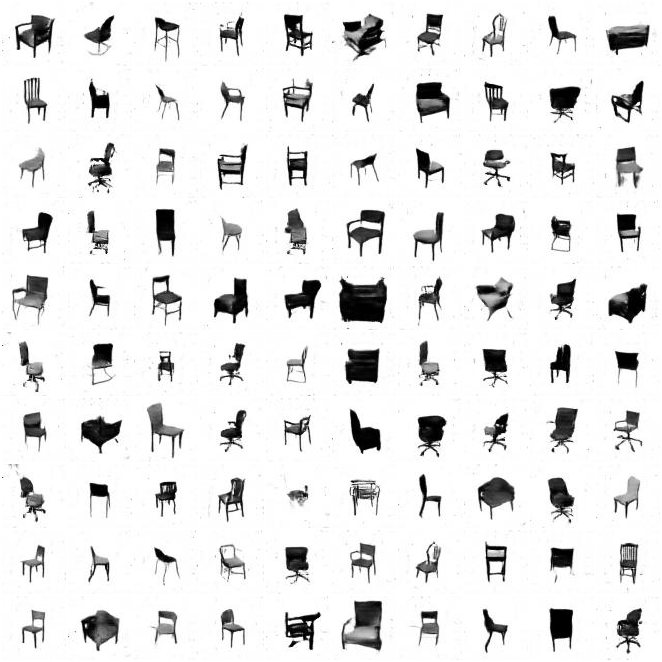} & 
\includegraphics[width=0.3\textwidth]{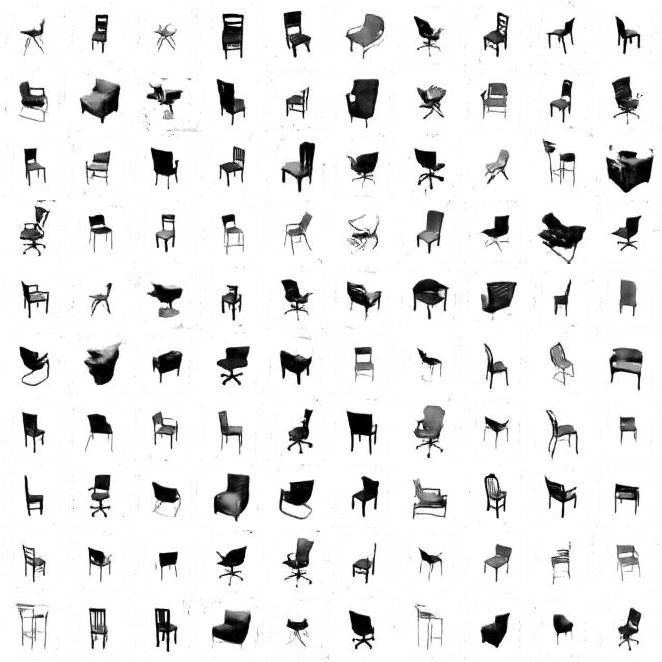} \\
(d) $\beta$-TCVAE  & (e) GAN & (f) InfoGAN\\[6pt]
\end{tabular}
\begin{tabular}{ccc}
\includegraphics[width=0.3\textwidth]{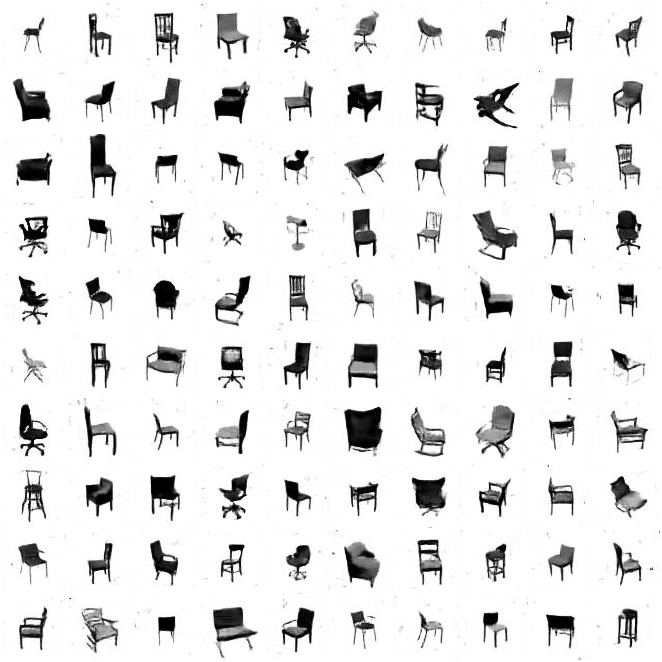} &
\includegraphics[width=0.3\textwidth]{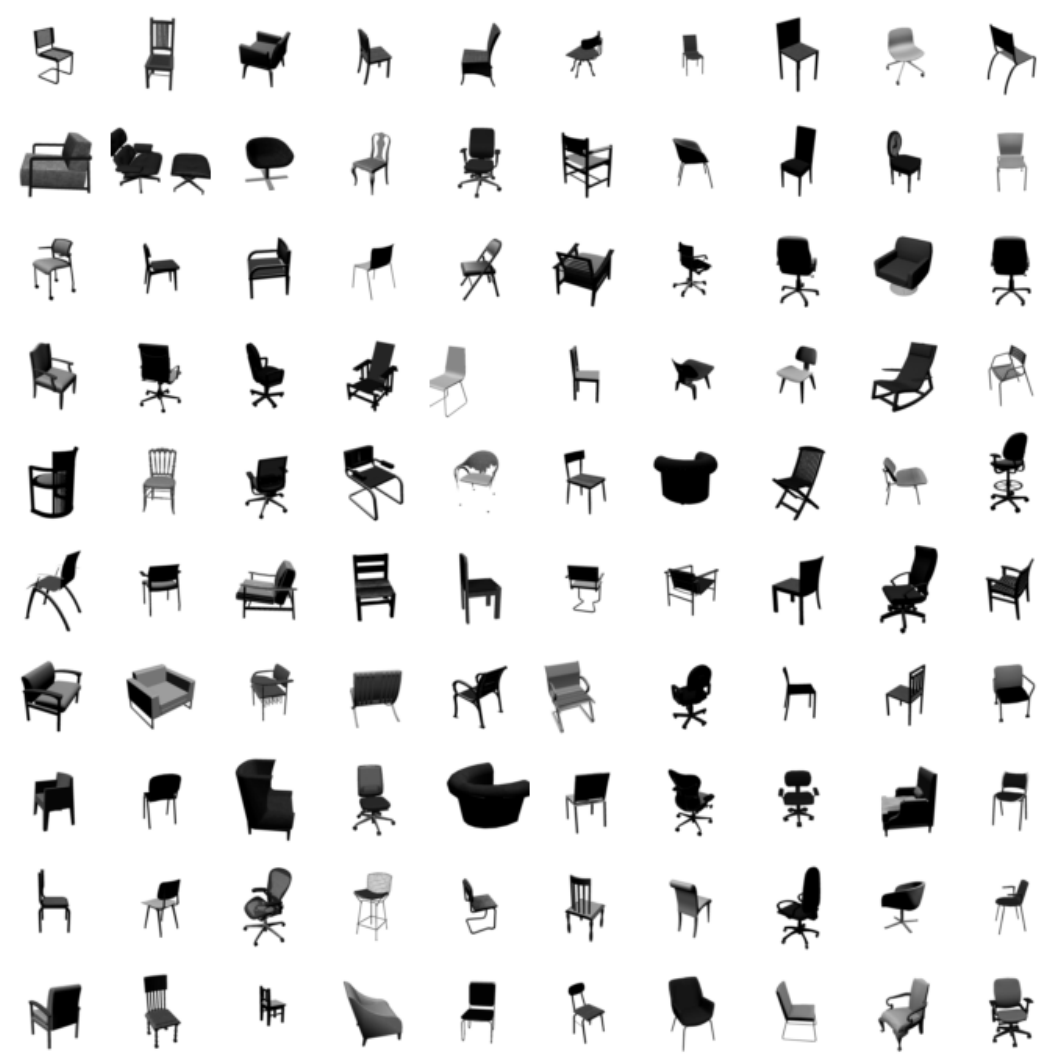} \\
(g) IB-GAN  & (h) Real Samples\\[6pt]
\end{tabular}
\caption{Comparison of randomly sampled images that are generated by the $\beta$-VAE and GAN baselines and IB-GAN on 3D Chairs dataset.}\label{fig:3dchairs_comp}
\end{figure*}

Figure \ref{fig:celeba_comp} and \ref{fig:3dchairs_comp} present randomly sampled images that are generated by the $\beta$-VAE and GAN baselines and IB-GAN on CelebA and 3D Chairs.
For fair comparison of generation performance, we use the same architecture for the decoders of VAE baselines with  the generator of IB-GAN.
For the encoders of $\beta$-VAE baselines, we reverse the architecture of the decoder networks. The generated images by IB-GAN are often sharper and more realistic than those of $\beta$-VAE baselines~\citep{Kingma:2013tz, Higgins:2016vm,Kim:2018th,Chen:2018wu}.


\clearpage



    
    


\section{C. ~~ Evaluation Metric}

\subsection{C.1 ~~ Disentanglement metric}

We employ the metric proposed by Kim \etal\cite{Kim:2018th} to evaluate the disentanglement performance of IB-GAN and other baselines. We use a batch of 100 samples to build each vote, and train a majority vote classifier using 800 votes. The accuracy of the classifier is then reported as the disentanglement performance for each model. Also, we exclude from consideration the collapsed latent dimensions of which empirical variances for the entire dataset are smaller than 0.05~\citep{Kim:2018th, google1}.

\subsection{C.2 ~~ FID score}

As commonly accepted, it is challenging to quantitatively evaluate how \textit{good} generative models are. 
Nonetheless, the FID score~\cite{heusel2017gans} is one possible candidate for measuring diversity and visual fidelity of generated samples. 
Precisely, the FID score measures the 2-Wasserstein distance between two distributions $p$ and $q$: $F(q,p) = || \mu_q -\mu_y||_2^2 + \mbox{trace}(C_q+C_p-2(C_q C_p)^{1/2})$, where $\{\mu_q,C_q\}$ and $\{\mu_p, C_p\}$ are respectively the mean and the covariance of the feature vectors produced by the inception model \cite{incep:43022} for true and generated samples. 
In both CelebA and 3D Chairs, we use 50,000 real and generated samples for the computation of $C_q$ and $C_p$, respectively.

\section{D. ~~ Model Architecture}


\begin{table*}[ht!]
    \centering
        \begin{tabular}{lcccc}
        \toprule
        Dataset & dSprites & Color-dSprites & CelebA & 3DChairs \\
        \midrule
        \ndf & $16$ & $16$ & $64$ & $32$ \\
        \midrule
        \ngf & $16$ & $16$ & $64$ & $32$ \\
        \bottomrule
        \end{tabular}
    \caption{The number of filters in the first layer of the generator (\ngf) and discriminator and (\ndf) for IB-GAN and GAN baselines.}
    \label{tab:ngf_ndf}
\end{table*}

\begin{table*}[ht!]  
    \centering
    \setlength{\tabcolsep}{9pt}
    \renewcommand{\arraystretch}{0.6}
        \begin{tabular}{ccc}
        \\
        \toprule
        Encoder (e) & Generator (G) & \makecell{Discriminator (D) /\\ Reconstructor (Q)} \TBstrut\\
        \midrule
        Input $z \in \mathbb{R}^{D_z}$ & 
        Input $r \in \mathbb{R}^{D_r}$ &
        Input $x \in \mathbb{R}^{64\times 64\times N_c}$\TBstrut\\
        \midrule
        \makecell{FC. $64$ ReLU.\\BN}&
        \makecell{FC. \ngf$\times 16$ ReLU.\\BN}&
        \makecell{$4\times 4$ conv. \texttt{ndf} lReLU.\\
        stride 2 (shared)}\TBstrut\\
        \midrule
        \makecell{FC. $64$ ReLU.\\BN}&
        \makecell{FC. $8\times 8\times$ \ngf$\times 4$ ReLU.\\BN}&
        \makecell{$4\times 4$ conv. \ndf$\times 2$ lReLU.\\
        stride 2. BN (shared)}\TBstrut\\
        \midrule
        \makecell{FC. $32$ ReLU.\\BN}&
        \makecell{$3\times 3$ upconv. \ngf$\times 4$ ReLU.\\
        stride 1. BN} &
        \makecell{$4\times 4$ conv. \ndf$\times 4$ lReLU.\\
        stride 2. BN (shared)}\TBstrut\\
        \midrule
        FC $D_r \times 2$ & 
        \makecell{$3\times 3$ upconv. \ngf$\times 4$ ReLU.\\
        stride 1. BN} &
        \makecell{$3\times 3$ conv. \ndf$\times 4$ lReLU.\\
        stride 1. BN}\TBstrut\\        
        \midrule
        \makecell{Reparametrization\\Trick}&
        \makecell{$4\times 4$ upconv. \ngf$\times 2$ ReLU.\\
        stride 2. BN} &
        \makecell{$3\times 3$ conv. \ndf$\times 4$ lReLU.\\
        stride 1. BN}\TBstrut\\ 
        \midrule
        --&
        \makecell{$4\times 4$ upconv. \ngf ReLU.\\
        stride 2. BN} &
        \makecell{$8\times 8$ conv. \ndf$\times 16$ lReLU.\\
        stride 1. BN}\TBstrut\\
        \midrule
        --& 
        \makecell{$4\times 4$ upconv. $N_c$ Tanh.\\
        stride 2} &
        \makecell{FC. $D_z$ for Q\\FC. $1$ for D}\TBstrut\\
        \bottomrule
        \\
        \end{tabular}
    \caption{The base architecture for IB-GAN on dSprites, CelebA, and 3DChairs. $N_c$ denotes the number of channels of input images, lReLU is the leaky relu activation and BN is the batch normalization. The layers shared by discriminator D and reconstructor Q are marked as shared.} 
    \label{tab:base_archi}
\end{table*}
\begin{table*}[ht!]  
    \centering
    \setlength{\tabcolsep}{9pt}
    \renewcommand{\arraystretch}{0.6}
        \begin{tabular}{ccc}
        \\
        \toprule
        Encoder (e) & Generator (G) & \makecell{Discriminator (D) /\\ Reconstructor (Q)} \TBstrut\\
        \midrule
        Input $z \in \mathbb{R}^{D_z}$ & 
        Input $r \in \mathbb{R}^{D_r}$ &
        Input $x \in \mathbb{R}^{64\times 64\times N_c}$\TBstrut\\
        \midrule
        \makecell{FC. $32$ ReLU.\\BN}&
        \makecell{FC. \ngf$\times 16$ ReLU.\\BN}&
        \makecell{$4\times 4$ conv. \texttt{ndf} lReLU.\\
        stride 2}\TBstrut\\
        \midrule
        \makecell{FC. $32$ ReLU.\\BN}&
        \makecell{FC. $8\times 8\times$ \ngf$\times 4$ ReLU.\\BN}&
        \makecell{$4\times 4$ conv. \ndf$\times 2$ lReLU.\\
        stride 2. BN}\TBstrut\\
        \midrule
        FC $D_r \times 2$ & 
        \makecell{$3\times 3$ upconv. \ngf$\times 4$ ReLU.\\
        stride 1. BN} &
        \makecell{$4\times 4$ conv. \ndf$\times 4$ lReLU.\\
        stride 2. BN}\TBstrut\\
        \midrule
        \makecell{Reparametrization\\Trick} & 
        \makecell{$3\times 3$ upconv. \ngf$\times 4$ ReLU.\\
        stride 1. BN} &
        \makecell{$3\times 3$ conv. \ndf$\times 4$ lReLU.\\
        stride 1. BN}\TBstrut\\        
        \midrule
        --&
        \makecell{$4\times 4$ upconv. \ngf$\times 2$ ReLU.\\
        stride 2. BN} &
        \makecell{$3\times 3$ conv. \ndf$\times 4$ lReLU.\\
        stride 1. BN}\TBstrut\\ 
        \midrule
        --&
        \makecell{$4\times 4$ upconv. \ngf ReLU.\\
        stride 2. BN} &
        \makecell{$8\times 8$ conv. \ndf$\times 16$ lReLU.\\
        stride 1. BN}\TBstrut\\
        \midrule
        --& 
        \makecell{$4\times 4$ upconv. $N_c$ Tanh.\\
        stride 2} &
        \makecell{FC. $D_z$ for Q\\FC. $1$ for D}\TBstrut\\
        \bottomrule
        \\
        \end{tabular}
    \caption{The base architecture for IB-GAN on Color-dSprites. $N_c$ denotes the number of channels of input images, lReLU is the leaky relu activation and BN is the batch normalization.} 
    \label{tab:base_archi_cdsprites}
\end{table*}

\newpage

\begin{table*}[ht!] 
    \centering
    \setlength{\tabcolsep}{9pt}
    \renewcommand{\arraystretch}{0.6}
        \begin{tabular}{cc}
        \\
        \toprule
        Generator (G) & \makecell{Discriminator (D) /\\ Reconstructor (Q)} \TBstrut\\
        \midrule
        Input $z \in \mathbb{R}^{D_z}$ &
        Input $x \in \mathbb{R}^{64\times 64\times N_c}$\TBstrut\\
        \midrule
        \makecell{FC. \ngf$\times 16$ ReLU.\\BN}&
        \makecell{$4\times 4$ conv. \texttt{ndf} lReLU.\\
        stride 2 (shared)}\TBstrut\\
        \midrule
        \makecell{FC. $8\times 8\times$ \ngf$\times 4$ ReLU.\\BN}&
        \makecell{$4\times 4$ conv. \ndf$\times 2$ lReLU.\\
        stride 2. BN (shared)}\TBstrut\\
        \midrule
        \makecell{$3\times 3$ upconv. \ngf$\times 4$ ReLU.\\
        stride 1. BN} &
        \makecell{$4\times 4$ conv. \ndf$\times 4$ lReLU.\\
        stride 2. BN (shared)}\TBstrut\\
        \midrule
        \makecell{$3\times 3$ upconv. \ngf$\times 4$ ReLU.\\
        stride 1. BN} &
        \makecell{$3\times 3$ conv. \ndf$\times 4$ lReLU.\\
        stride 1. BN}\TBstrut\\        
        \midrule
        \makecell{$4\times 4$ upconv. \ngf$\times 2$ ReLU.\\
        stride 2. BN} &
        \makecell{$3\times 3$ conv. \ndf$\times 4$ lReLU.\\
        stride 1. BN}\TBstrut\\ 
        \midrule
        \makecell{$4\times 4$ upconv. \ngf ReLU.\\
        stride 2. BN} &
        \makecell{$8\times 8$ conv. \ndf$\times 16$ lReLU.\\
        stride 1. BN}\TBstrut\\
        \midrule
        \makecell{$4\times 4$ upconv. $N_c$ Tanh.\\
        stride 2} &
        \makecell{FC. $D_z$ for Q\\FC. $1$ for D}\TBstrut\\
        \bottomrule
        \\
        \end{tabular}
    \caption{The base architecture for InfoGAN. This architecture is shared in the experiments on dSprites and Color-dSprites. $N_c$ denotes the number of channels of input images, lReLU is the leaky relu activation and BN is the batch normalization. The layers shared by discriminator D and reconstructor Q are marked as shared. All hyperparameters are the same as those of IB-GAN except $D_z=10$.}
    \label{tab:base_archi_infogan}
\end{table*} 


\clearpage

\section{E. ~~ Implementation Details}
\label{sec:appendix_implementation} 

\begin{table*}[ht!] 
    \centering
    \setlength{\tabcolsep}{4pt}
    \renewcommand{\arraystretch}{1.5}
        \begin{tabular}{cccccc} 
        \\
        \toprule
          Dataset & Hyperparameters & Learning rates & Iterations & Instance noise & Label smoothing \\
        \midrule
        \multirow{2}{*}{dSprites} & $D_z$=$16$, $D_r$=$10$ &  G/E/Q: 5e-5, & \multirow{2}{*}{1.5e5} & 1 $\rightarrow$ 0 & \multirow{2}{*}{No} \\ 
        & $\gamma$=1, $\beta$=0.141 & D: 1e-6 & & for 1e5 iters & \\ 
        \midrule
        \multirow{2}{*}{Color-dSprites} & $D_z$=$16$, $D_r$=$10$ &  G/E/Q: 5e-5, & \multirow{2}{*}{5e5} & 1 $\rightarrow$ 0 & \multirow{2}{*}{No} \\ 
        & $\gamma$=1, $\beta$=0.071 & D: 1e-6 & & for 4e5 iters & \\ 
        \midrule
        \multirow{2}{*}{3D Chairs} & $D_z$=$64$, $D_r$=$10$ &  G/E/Q: 5e-5, & \multirow{2}{*}{7e5} & 0.5 $\rightarrow$ 0.01 & \multirow{2}{*}{Yes} \\ 
        & $\gamma$=1.2, $\beta$=0.35 & D: 2e-6 & & for 7e5 iters & \\ 
        \midrule
        \multirow{2}{*}{CelebA} & $D_z$=$64$, $D_r$=$15$ &  G/E/Q: 5e-5, & \multirow{2}{*}{1e6} & 0.5 $\rightarrow$ 0.01 & \multirow{2}{*}{Yes} \\ 
        & $\gamma$=2, $\beta$=0.325 & D: 2e-6 & & for 1e6 iters & \\ 
        \bottomrule
        \\
        \end{tabular}
    \vspace{-12pt}
  \caption{The hyperparameter settings for IB-GAN in all experiments. We use RMSProp with momention of 0.9.
    In hyperparameters, $D_z$ and $D_r$ mean the dimension of $z$ and $r$. 
    In learning rates, G,E,Q,D indicates generator, encoder, reconstructor and discriminator.
    In instance noise, $\sigma_{inst}$ is annealed linearly between the two values for the following iterations.}
    \label{tab:hyperparams}
\end{table*}

\textbf{Stabilization of GAN training}.
The training of GANs is notoriously unstable~\citep{are46506, pmlr-v80-mescheder18a}. 
To stabilize the training of GAN-based models in our experiments, we adopt two popular tricks: the instance noise technique~\citep{Sonderby2016a} and one-sided label smoothing~\citep{salimans2016improved}. For the instance noise technique, we add instance noises $\epsilon \sim N(0, \sigma_{inst}*I)$ to both real and generated images while linearly decreasing the value of $\sigma_{inst}$ during training iterations. For the one-sided label smoothing technique, we sample true labels from a uniform distribution within the range of [0.7, 1.2]. 
Table \ref{tab:hyperparams} summarizes important hyper-parameters including these two stabilization regularizers.

\section{F. ~~ Datasets}
\label{sec:dataset_details}



\begin{table*}[ht!] 
    \centering
    \setlength{\tabcolsep}{3pt}
    \renewcommand{\arraystretch}{1}
        \begin{tabular}{cc} 
        \\
        \toprule
        Dataset & Specification \\
        \midrule
        \multirow{4}{*}{dSprites~\citep{Higgins:2016vm}} & \multirow{4}{25em}{737,280 binary $64\times64$ images of 2D shapes with 5 ground-truth factors, which consist of 3 shapes, 6 scales, 40 orientations and 32 positions of $X$ and $Y$.} \\
        & \\
        & \\
        & \\
        \midrule
        \multirow{7}{*}{\cite{Burgess:2018uf, google1}} & \multirow{6}{25em}{RGB $64\times64\times3$ images of 2D shapes with 6 ground truth factors. All factors are identical to those of dSprites dataset, except for an additional \textit{color} factor; it is quantized into 256 different bins, which are obtained by discretizing each color channel into 8 values linearly spaced between [0, 1].} \\
        & \\
        Color-dSprites & \\
        & \\
        & \\
        & \\
        \midrule
        \multirow{3}{*}{3D Chairs~\citep{Aubry14}} & \multirow{3}{25em}{86,366 gray-scale $64\times64$ images of 1,393 chair CAD models with 31 azimuth angles and 2 elevation angles.} \\
        & \\
        & \\
        \midrule
        \multirow{4}{*}{CelebA~\citep{liu2015faceattributes}} & \multirow{4}{25em}{202,599 RGB $64\times64\times3$ images of celebrity faces consisting of 10,177 identities, 5 landmark locations and 40 binary attributes. We use the cropped version of the dataset.} \\
        & \\
        & \\
        & \\
        \bottomrule
        \\
        \end{tabular}
    \vspace{-9pt}
    \caption{The specification of datasets.}
    \label{tab:dataset_spec}
\end{table*}

\clearpage

\end{document}